\documentclass[acmtog]{acmart}
\acmSubmissionID{}

\usepackage{amsbsy}
\usepackage[mathscr]{euscript}
\usepackage{amsmath}
\usepackage{stackrel}
\usepackage{nicefrac}
\usepackage{booktabs}
\usepackage{placeins}

\usepackage{algpseudocode}
\usepackage{algorithm}
\usepackage{dblfloatfix}
\usepackage{subcaption}

\usepackage{float}

\usepackage{tikz}
\usepackage{xfrac}
\usepackage{balance}

\setcitestyle{square}

\acmJournal{TOG}

\graphicspath{{figures//}}

\definecolor{Gray}{rgb}{0.5,0.5,0.5}
\definecolor{darkblue}{rgb}{0,0,0.7}
\definecolor{blue}{rgb}{0,0,1}
\definecolor{orange}{rgb}{1,.5,0} 
\definecolor{red}{rgb}{1,0,0} 
\definecolor{black}{rgb}{0,0,0}
\definecolor{green}{rgb}{0,0.8,0}

\newcommand{\changed}[1]{{\textcolor{black}{#1}}}
\newcommand{\revised}[1]{{\textcolor{black}{#1}}}
\newcommand{\final}[1]{{\textcolor{black}{#1}}}

\newcommand{\ra}[1]{\renewcommand{\arraystretch}{#1}}

\newcommand{\mat}[1]{\mathbf{#1}}

\newcommand{\psf}{k}

\newcommand{\cfa}{\mat{S}}
\newcommand{\sys}{\mat{A}}
\newcommand{\cc}{\mat{E}}

\newcommand{\argmin}[1]{\stackrel[#1]{}{\textrm{argmin}}}
\newcommand{\minimize}[1]{\stackrel[#1]{}{\textrm{min}}}

\newcommand{\BEAS}{\begin{eqnarray*}}
\newcommand{\EEAS}{\end{eqnarray*}}
\newcommand{\BEA}{\begin{eqnarray}}
\newcommand{\EEA}{\end{eqnarray}}
\newcommand{\BEQ}{\begin{equation}}
\newcommand{\EEQ}{\end{equation}}
\newcommand{\BIT}{\begin{itemize}}
\newcommand{\EIT}{\end{itemize}}
\newcommand{\BNUM}{\begin{enumerate}}
\newcommand{\ENUM}{\end{enumerate}}

\newcommand{\BA}{\begin{array}}
\newcommand{\EA}{\end{array}}

\newcommand{\eg}{{\it e.g.}}
\newcommand{\ie}{{\it i.e.}}

\newcommand{\ones}{\mathbf 1}

\newcommand{\Argmin}{\mathop{\rm argmin}}
\newcommand{\Argmax}{\mathop{\rm argmax}}
\newcommand{\prox}[1]{\mathbf{prox}_{#1}}

\newcommand{\higherloss}{\mathcal{L}}
\newcommand{\lowerloss}{\mathcal{G}}

\newcommand{\var}{\nu}

\newcommand{\ansc}{\mathcal{A}}

\title{Dirty Pixels: Towards End-to-End Image Processing and Perception} 

\begin{document}
\author{Steven Diamond}
\authornote{\final{These authors contributed equally to this research.}}
\affiliation{%
  \institution{Stanford University}
  }
\email{diamond@cs.stanford.edu}

\author{Vincent Sitzmann}
\authornotemark[1]
\affiliation{%
  \institution{Stanford University, MIT}
  }
\email{sitzmann@mit.edu}

\author{Frank Julca-Aguilar}
\authornotemark[1]
\affiliation{%
  \institution{Algolux}
  }
\email{frank.julca-aguilar@algolux.com}

\author{Stephen Boyd}
\affiliation{%
  \institution{Stanford University}
  }
\email{boyd@stanford.edu}

\author{Gordon Wetzstein}
\affiliation{%
  \institution{Stanford University}
  }
\email{gordonwz@stanford.edu}

\author{Felix Heide}
\affiliation{%
  \institution{Princeton University}
  }
\email{fheide@cs.princeton.edu}

\begin{teaserfigure}
	\centering
	\vspace{-4pt}
	\includegraphics[width=\textwidth]{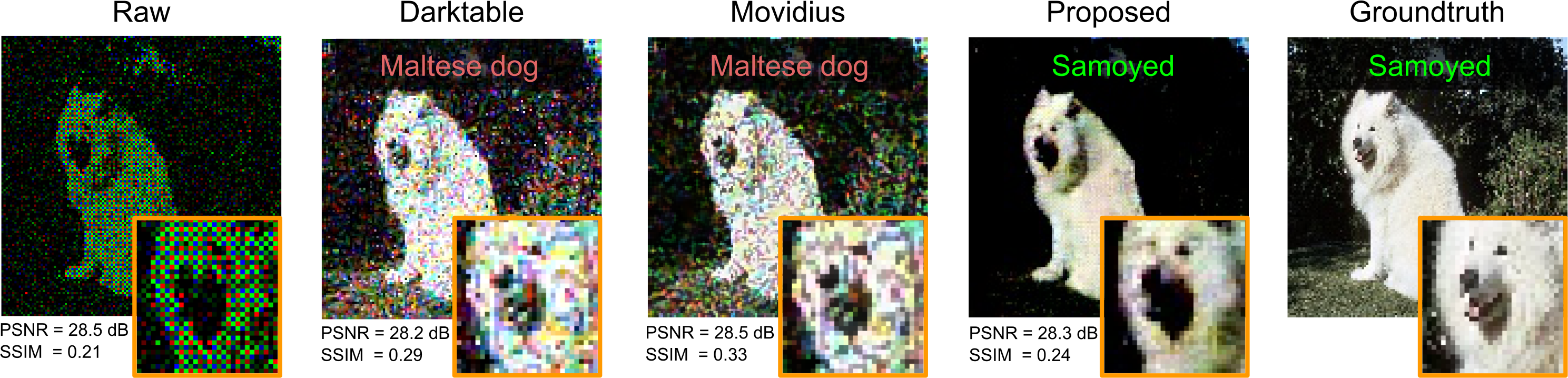}\vspace{-6pt}
   \caption{A RAW input image subsampled on a color filter array and corrupted by sensor characteristics in low light (left) and
   its class prediction using MobileNet-v1 along with conventional processing pipelines. 
	Processing RAW data using conventional image processing pipelines (ISPs)
 does not necessarily improve performance because conventional pipelines are
     optimized for human viewing, not for machine vision. Here, the image of a \changed{Samoyed} dog is missclassified as the much smaller Maltese dog with thinner coat and smaller snout. 
     We propose an end-to-end architecture for joint demosaicking, denoising,
     deblurring, and classification that makes classification robust in low-light scenarios.
    The proposed architecture learns a processing pipeline optimized for classification, which
    enhances fine details relevant for this high-level task -- at the expense 
    of more noise as measured by conventional metrics, PSNR and SSIM -- and improves state-of-the art accuracy. Here, the dog's snout, ears, fur and outline are enhanced in contrast at the loss of surrounding background class regions.
The proposed architecture has a principled and modular design and generalizes across light levels and cameras.
    }
		\vspace{4pt}
   \label{fig:teaser}
\end{teaserfigure}

\begin{abstract}
Real-world imaging systems acquire measurements that are degraded by noise, optical 
aberrations, and other imperfections that make image processing for human viewing 
and higher-level perception tasks challenging. Conventional cameras address this 
problem by \changed{compartmentalizing imaging from high-level task processing.
As such, conventional imaging involves processing the RAW sensor measurements in a sequential 
pipeline of steps, such as demosaicking, denoising, deblurring, 
tone-mapping and compression. This pipeline is optimized to obtain a visually pleasing 
image. High-level processing, on the other hand, involves steps such as feature extraction, 
classification, tracking, and fusion.} 
While this silo-ed design approach allows for efficient development, it also dictates 
compartmentalized performance metrics, without knowledge of the higher-level task 
of the camera system. For example, today's demosaicking and denoising algorithms 
are designed using perceptual image quality metrics but not with domain-specific 
tasks such as object detection in mind. We propose an 
end-to-end differentiable architecture that jointly performs demosaicking, denoising, 
deblurring, tone-mapping, and classification. The architecture 
\changed{does not require any intermediate losses based on perceived image quality} and  
learns processing pipelines whose outputs differ from those of existing ISPs 
optimized for perceptual quality, preserving fine detail at the cost of 
increased noise and artifacts. \changed{We show} that state-of-the-art ISPs 
discard information \changed{that} is essential in corner cases, such as extremely low-light 
conditions, where conventional imaging and perception stacks fail. 
We demonstrate on captured and simulated data that \changed{our} model substantially 
improves perception in low light and other challenging conditions, 
which is imperative for real-world applications like autonomous 
driving, robotics, and surveillance. 
\changed{Finally, we found that the proposed model also achieves state-of-the-art accuracy 
when optimized for image reconstruction in low-light conditions, validating the 
architecture itself as a potentially useful drop-in network for reconstruction and 
analysis tasks beyond the 
applications demonstrated in this work. 
Our proposed models, datasets, and calibration data are available at 
\url{https://github.com/princeton-computational-imaging/DirtyPixels}}

\end{abstract}

%
%



\setcopyright{acmcopyright}
\acmJournal{TOG}
\acmYear{2021} \acmVolume{1} \acmNumber{1} \acmArticle{1} \acmMonth{1}
\acmPrice{15.00}\acmDOI{10.1145/3446918}

\maketitle


\section{Introduction}
\label{sec:intro}
Image sensor measurements are affected by various degradations in the physical image formation process. Raw sensor readings suffer from photon shot noise, optical aberration, read-out noise, spatial subsampling in the color filter array (CFA), spectral cross-talk on the CFA, motion blur, and other imperfections. The image signal processor (ISP) is a hardware block that addresses these degradations by processing the RAW measurement in a sequential pipeline of steps~\cite{ramanath2005color} each targeting a sub-problem in isolation, before displaying or saving the resulting output image. The ISP performs an extensive set of operations, such as demosaicking~\cite{zhang2011color}, denoising, deblurring, and tone-mapping. 
All of these low-level imaging tasks are ill-posed problems with recent active research~\cite{gharbi2016deep,zhang2016beyond,heide2014flexisp,chen2018learning}. 
Existing image reconstruction algorithms are designed to minimize an explicit or implicit reconstruction loss aligned with human perceptions of image quality, as a prior to resolve the ill-posedness of the sub-problems listed above. Explicit losses are based on chart-based metrics~\cite{phillips2018camera}, and emerging domain-specific standards, such as CPIQ~\cite{jin2017towards}, DxOMark, VCX Score for cellphone imaging and the emerging IEEE~P2020~standard~\cite{p2020} for autonomous vehicles. However, the approach widely adopted by ISP manufacturers is to design and tune ISPs to eliminate artifacts \emph{human experts find visually unpleasant}, thereby minimizing an implicit perceptual loss. 

At the same time, applications in emerging domains, including autonomous driving, robotics, and surveillance, consume images directly by a higher-level analysis module without ever being viewed by humans. Human expert assessment is not applicable to these ``image-free'' cameras, and this gives rise to the question if low-level processing is necessary, or if existing higher-level networks should better be trained directly on RAW sensor data. 

ISPs are useful in that they map data from diverse camera systems into a common interface, a visually pleasing image, that most large-scale computer vision datasets adopt, \eg, \cite{deng2009imagenet,lin2014microsoft}. For downstream tasks, the real-world performance of a deployed high-level network will be close to the performance on clean images so long as the low-level pipeline can approximately recover the latent clean image from RAW data. However, in challenging capture conditions, \ie, the corner cases of the ISP, recovering the latent image is extremely challenging, such as low-light captures that are heavily degraded by photon shot noise. For example, a denoising block that is optimized for perceptual quality will remove apparent chromatic noise, \eg, the Movidius Myriad 2 ISP includes a Chroma-NLM stage for perceptual quality~\cite{moloney2014myriad}, thereby destroying high-frequency color detail that could be exploited in the higher-level image analysis. Identical design trade-offs are found other key processing blocks, such as demosaicking, tone-mapping, and sharpening~\cite{moloney2014myriad}.

An immediate solution for such failure modes appears to be removing the ISP completely and training the perception model directly on RAW measurement data. That way no information will be suppressed in the low-level image processing modules. Indeed, we demonstrate that existing classifiers trained on RAW data perform on-par with pre-processing from traditional ISPs, hand-crafted for perceptual viewing instead of CNN feature extraction.

In this work, we depart from traditional ISPs, and investigate learned architectures that perform end-to-end image processing and classification jointly. \changed{We propose an end-to-end differentiable model} that uses RAW color filter array data as input and outperforms existing deep classification directly trained \changed{on this} RAW input streams by a more than 5\% in top-5 accuracy on in-the-wild captures. 
We validate that low light is indeed a failure mode for conventional computer 
vision systems that combine existing ISPs with existing high-level networks. 
We propose a novel neural architecture for joint denoising and demosaicking, 
dubbed ``Anscombe networks'', that we learn jointly with a high-level network 
and that exploits knowledge of the camera image formation model. We show that 
fine-tuning an Anscombe network with a high-level model performs better 
than training a high-level model directly on the RAW data or on the output 
of traditional ISPs, \changed{or recent 
state-of-the-art learnable ISP~\cite{chen2018learning}}. We demonstrate that the proposed 
Anscombe network ISP generalizes across imaging setting akin to a traditional ISP. 
Nevertheless, the output of the neural ISP differs from that of traditional 
ISPs, scoring worse on traditional perceptual metrics when trained for classification. 
However, when trained \final{for human viewing}, 
and no downstream analytic task, the proposed architecture achieves 
state-of-the-art image quality for low-light imaging, \changed{highlighting the potential of domain-specific imaging pipelines.}

The contributions of this paper are the following:
\begin{itemize}
\item We demonstrate that conventional perception pipelines, which use a state-of-the-art ISP and classifier trained on a standard JPEG dataset, perform poorly in low light.
\item We introduce Anscombe networks, a \changed{light-weight} neural camera ISP for demosaicking and denoising that generalizes across camera architecture and capture settings. We show that Anscombe networks, by themselves, achieve state-of-the-art image quality when trained for low-light imaging using a perceptual loss for image quality.
\item We demonstrate that jointly learning Anscombe networks with classification networks outperform training the high-level networks directly on RAW data or the output of state-of-the-art software, hardware \changed{and learnable} ISPs, both when trained from scratch or fine-tuned.
\item We evaluate the joint end-to-end model on synthetic and captured RAW data. To this end, we introduce a dataset of realistic noise and blur models calibrated from mobile cameras and a dataset of annotated noisy RAW captures.
\item We demonstrate a real-time smart-phone implementation of the proposed end-to-end low-light classification model.
\end{itemize}

\changed{
In the future, a large portion of our images will be consumed by high-level perception stacks, not by humans. 
We propose to reexamine the foundational assumptions of image processing (ISPs). 
Existing approaches tackle this challenge either by discarding ISPs and retraining 
downstream networks directly on RAW data, or they manually tune, or optimize 
the parameters of hardware ISPs for a fixed network. 
Our work departs from both approaches and, to the best of our knowledge, it is 
the \emph{first that jointly learns image processing (ISP) and classification network parameters} 
in an end-to-end fashion. We note that \emph{specialized domain-specific 
processing is the goal of the proposed approach}. We do not dismiss traditional 
ISPs for general imaging tasks with unknown downstream applications but illustrate 
the potential of domain-specific camera processing.
}

%

\section{Related Work}
\label{sec:related}
\paragraph{Effects of Noise and Blur on High-level Networks}

A small body of work has explored the effects of noise and blur on deep networks trained for high-level vision tasks.
Dodge and Karam evaluated a variety of recent classification networks under noise and blur and found a substantial drop in performance \shortcite{dodge2016understanding}.
Vasiljevic et al.~similarly showed that blur decreased classification and segmentation performance for deep nets,
though much of the lost performance was regained by fine-tuning on blurry images \shortcite{vasiljevic2016examining}.
Karahan et al.~showed that noise and blur degrade the performance of CNNs trained for face recognition \shortcite{karahan:16}.
Several authors demonstrated that preprocessing noisy images with trained or classical denoisers
improves the downstream performance \cite{TE:10,tang2012robust,burges:13,JNWM:16,da2016empirical}.
Chen et al.~showed that training a model for denoising and separately classification 
can improve performance on both tasks \shortcite{chen2016joint} 
\changed{when tested on corrupted versions of the MNIST and USPS datasets.} 
\changed{Note that the models \emph{trained from scratch}, in Tables~\ref{tab:benchmark} and~\ref{tab:raw_benchmark},
are equivalent to Chen et al.~\shortcite{chen2016joint} approach, where we optimize the 
classification network directly from RAW data}.

\paragraph{Camera Image Processing Pipelines}
Most digital cameras perform low-level image processing such as denoising and demosaicking in a hardware ISP pipelines based on efficient heuristics~\cite{ram:05,zhang2011color,shao:14}.
Modern imaging systems for cellphone use-cases may acquire a burst of images or images from multiple camera modules.
Recently, Hasinoff et al.~\shortcite{hasinoff2016} have demonstrated high-quality imaging in low light using bursts, which are then processed in a software ISP tuned for perceptual quality.
Cameras for driver assistant systems, autonomous cars or other robotic purposes, however, have to react in real-time and therefore cannot acquire sequential exposures, leading to the emerge of split-pixel sensors (OmniVision OV10640, OV10650) and domain specific ISPs, such as the ARM Mali C71.
Most conventional camera ISPs are implemented as fixed-function ASIC blocks to handle high-resolution image feeds at real-time rates~\cite{aptinaMT9P111}.
Only recently, camera ISPs are starting to become more programmable
\changed{(also the case for software ISPs such as Hasinoff et al.~\shortcite{hasinoff2016})}. 
The Movidius Myriad 2~\cite{moloney2014myriad} hardware ISP offers configurable pipelines 
with room for a few general-purpose blocks run on SIMD Vector Processors, 
but still relies on a large number of fixed-function hardware blocks.
Hegarty et al.~\shortcite{Hegarty:2014} propose a domain-specific language for camera ISP processing on FPGAs,
which translates image processing pipelines into efficient, low-power FPGA architectures.
Instead of designing pipelines, Heide et al.~\shortcite{heide2014flexisp} pose low-level image processing as an optimization problem, achieving higher quality than previous ISPs for a variety of camera systems.
However, their iterative optimization method is computationally intensive and an order of magnitude slower than real-time.
Recently, Gharbi et al.~rely on deep convolutional architectures to perform low-level vision tasks, such as demosaicking~\shortcite{gharbi2016deep} or tonemapping~\shortcite{GharbiSIGGRAPH2017}.
While being computationally efficient, their architectures \changed{depend on heavily engineered datasets for training their models, whereas we use standard classification datasets}. 
\final{Liba et al.~\shortcite{Liba:2019} proposed a system for capturing 
images in low-light conditions based on the alignment and combination 
of multiple frames, and learning-based white balance and tonemapping.}
\changed{Schwartz et al.~\shortcite{Liang:2019}, Liang et al.~\shortcite{Schwartz:2019},  and 
Chen et al.~\shortcite{chen2018learning} proposed learnable ISPs based on 
deep convolutional networks}. \changed{The model proposed by  
Chen et al.~\shortcite{chen2018learning} consists of convolutional network with CFA pixel packing 
similar to~\cite{gharbi2016deep}}. While their results are perceptually 
on-par or better than naive post-filtering approaches, using 
BM3D~\cite{dabov2007image} as an artifact suppression block, it remains unclear 
if recent state-of-the-art ISPs using traditional denoising blocks on RAW 
data, i.e., not as post-processing artifact suppression block, 
perform better as concluded in~\cite{Plotz_2017_CVPR}. 
\changed{Our results described in Section~\ref{sec:assessment} 
show that our proposed Anscombe ISP improves accuracy 
of a classifier trained on top of Chen et al.~\shortcite{chen2018learning}-preprocessed (and finetuned) images.}

\paragraph{Traditional Image Processing Pipelines for Computer Vision}
The role of traditional hardware ISP components in vision systems was examined in \cite{Buckler2017,tseng2019hyperparameter,yahiaoui2019optimization}. Buckler 
et al.~\shortcite{Buckler2017} suggested that ISPs should be configurable to switch 
between a human-viewable mode and computer vision mode to produce data 
optimized for vision tasks. However, ISP parameter tuning by visual inspection 
is extremely challenging if performed manually, motivating simulation environments~\cite{blasinski2018optimizing}. Simulated environments, unfortunately, 
suffer from a significant domain gap~\cite{hoffman2017cycada}. Recently 
Tseng et al.~\shortcite{tseng2019hyperparameter} proposed an automatic method 
for optimizing black-box ISPs. They propose to model and learn a differentiable 
proxy function that approximates the entire image processing pipeline. 
\changed{In contrast to the proposed method, Tseng et al. rely on traditional hardware ISPs and optimize \emph{only 
ISP hyperparameters, not the high-level network}. The efficacy of this
approach relies on the accuracy of the ISP approximation. As such, in our low-light scenario, the approximator network from Tseng et al. failed, see Figure 5 Supplement.
We note that none of the above methods propose a \emph{jointly end-to-end optimized ISP and downstream network}.} 
%

\paragraph{Domain Adaptation}
A common problem in deep neural networks trained for high-level computer vision tasks is domain shift,
meaning the difference in image statistics between the training data and the unknown real-world data,
leading to poor performance of a trained model in the final real-world scenario.
The literature on domain adaptation includes many methods for adapting models trained on one distribution to a target distribution, ranging in sophistication from simply fine-tuning the model on 
labeled data from the target distribution to more recent work that only requires sparsely labeled or unlabeled data
(\eg, \cite{tzeng2017adversarial,long2015learning,ganin2015unsupervised,sun2016deep,tzeng2014deep}).
The domain adaptation literature has an implicit assumption, however,
that the mapping from the training domain to the target domain is unknown.
In our problem of classification under noise and blur,
the mapping from the clean training data to degraded real world data can be modeled extremely accurately.
We are thus able to map the clean training data into the target domain through simulation,
and moreover to efficiently incorporate our a-priori knowledge of the physical model into the classification architecture.
\changed{We put this into practice in the design of our efficient Anscombe networks, which, as 
validated in Section~\ref{sec:results}, outperform existing image processing layers when integrated and end-to-end trained.}

\section{Camera Image Formation Model}
\label{sec:calibration}
\subsection{Image formation}\label{sec:image_formation}
\begin{figure}[t!]
	\centering
		\includegraphics[width=\columnwidth]{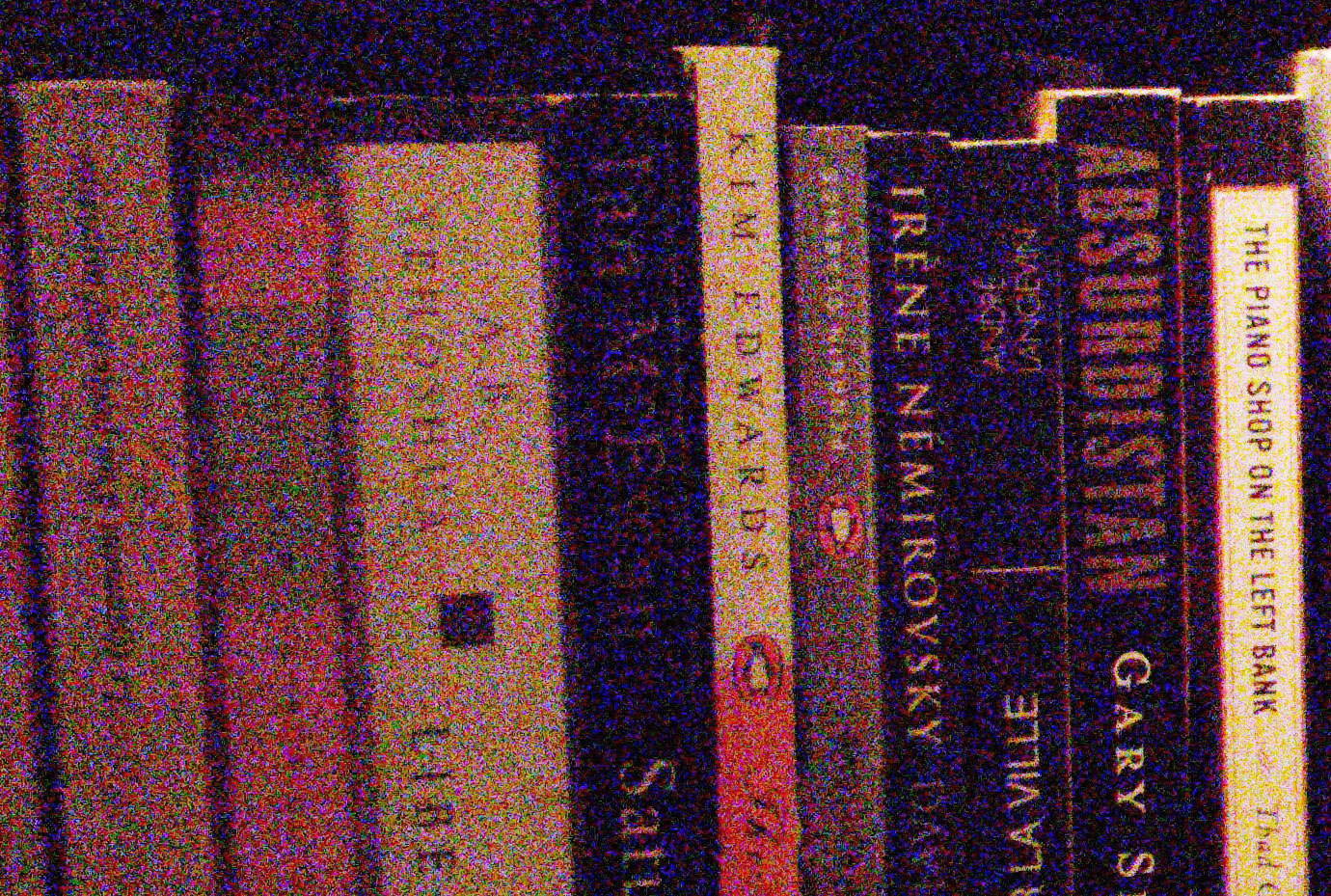}
		\caption{A RAW frame captured indoors using a Nexus 6 rear camera (after demosaicking).
      The image was taken at ISO 3000 with a 32 ms exposure time.
      The noise in the image is clearly visible.}
		\label{fig:nexus_capture}
\end{figure}
We consider the image formation \changed{$\mathcal{I}$} for a RAW sensor image as
\begin{equation}\label{eq:imaging_PSF}
\begin{aligned}
  y_x &\sim \alpha\mathcal{P} \left( \sum_{c \in \{R,G,B\}} \cfa^c (\psf_c \ast \cc_c x)/\alpha \right)  + \mathcal{N}(0,\sigma^2) \\
	\Leftrightarrow y_x & \sim  \alpha\mathcal{P} \left( \sys x/\alpha \right)  + \mathcal{N}(0,\sigma^2) \\
  y &= \changed{\mathcal{I}}(x) = \Pi_{[0,1]}(y_x),
\end{aligned}
\end{equation}
where $x \in \mathbb{R}^{3N}$ is the vectorized latent color image, with $N$ being the number of pixels, $y \in \mathbb{R}^{N}$ is the measured RAW image, $\alpha >
0$ and $\sigma > 0$ are parameters in a Poisson and Gaussian distribution, respectively, the operator $\cc_c$ extracts the color channel $c \in \{R,G,B\}$, $\psf_c$
represents the lens point spread function (PSF) in the color channel $c$,
$\ast$ denotes the linear operator corresponding to 2D convolution on the vectorized input, and $\Pi_{[0,1]}$
denotes projection onto the interval $[0,1]$. The matrix $\cfa^c$ models the spatial sub-sampling for color filter $c$ on the color filter array of the sensor. This matrix is a diagonal sub-sampling matrix defined as
\begin{equation}
\label{eq:subsample_matrix}
\cfa^c_{ii} = \left\{ \begin{array}{l}
1    \quad \text{if pixel $t$ has color filter $i$},\\
0   \quad  \text{else},
\end{array} \right.
\end{equation}
\begin{figure*}[t!]
	\centering
		\includegraphics[width=\textwidth]{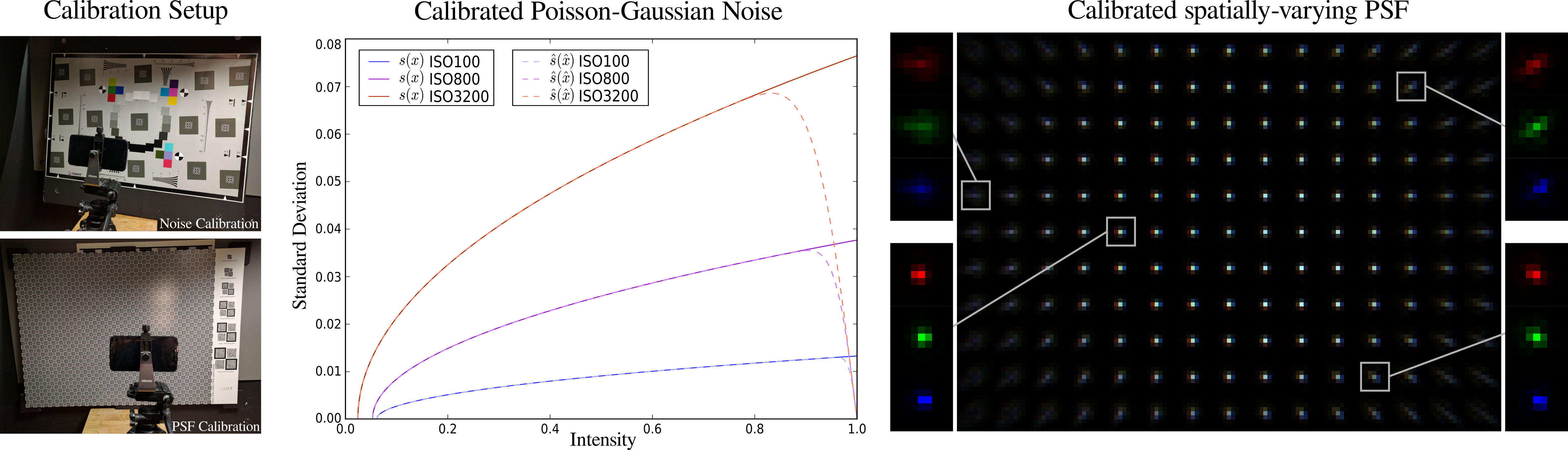}
		\caption{(Top left) The noise calibration setup. (Bottom left) The PSF calibration setup. (Center) $s(x) = \mathrm{std}(\tilde{y})$
versus $E[\tilde{y}]$ and $\hat{s}(\hat{x}) = \mathrm{std}(y)$ versus $E[y]$
for different ISO levels on a Nexus 6P rear camera.
The noise parameters $\alpha$ and $\sigma$ at a given light level are computed
from the $s(x)$ and $\hat{s}(\hat{x})$ plots. (Right) The PSFs for the entire field-of-view of a Nexus 5 rear
      camera. Two center PSFs, an off-axis PSF, and a periphery PSF are
      magnified.
     }
		\label{fig:calibration}
\end{figure*}
The image formation model from above is composed of a linear part $\sys x$, modeling all optical effects in the capture process with the matrix $\sys$, and a non-linear sampling process according to the noise characteristics of the sensor. The measured image follows the physically accurate Poisson-Gaussian noise
model with clipping described by Foi et al.~\shortcite{foi08,foi2009clipped}.
In the noise model, decreasing the light level increases $\alpha$,
but the dynamic range is kept constant by increasing the ISO,
represented by multiplying $\mathcal{P}(\sys x/\alpha)$ by $\alpha$. 

The image formation model from Eq.~\eqref{eq:imaging_PSF} is general and applicable to a variety of different camera architectures, ranging from traditional Bayer CFA cameras to interlaced HDR sensors, each covered by changing the linear forward model $\sys$ according to the given camera architecture. We refer the reader to~\cite{heide2014flexisp} for a variety of camera architectures this model supports. 
Note that, in contrast to~\cite{gharbi2016deep,heide2014flexisp}, we assume a more accurate noise model, including the Poissonian component which is critical for the model accuracy in the low-flux regime. 

\subsection{Calibration}
\label{sec:calibration}
We calibrated the parameters $\alpha$, and $\sigma$ of the image
formation model from Sec.~\ref{sec:image_formation} 
by acquiring calibration captures of a charts containing patches of different shades
of gray (\eg, \cite{iso12233}) at various gains with auto-white-balance disabled. We then \changed{follow Foi et al.~\shortcite{foi2009clipped}} to estimate the unknown noise parameters.
The photograph on the left in Fig.~\ref{fig:calibration} shows our noise calibration setup.
The center plot in Fig.~\ref{fig:calibration} shows plots of $s(x) = \mathrm{std}(\tilde{y})$
versus $E[\tilde{y}]$ and $\hat{s}(\hat{x}) = \mathrm{std}(y)$ versus $E[y]$
for different ISO levels on a Nexus 6P rear camera. 
The parameters $\alpha$ and $\sigma$ at a given light level are computed from
the $s(x)$ and $\hat{s}(\hat{x})$ plots.
The noise under our calibrated image formation model can be high. Fig.~\ref{fig:nexus_capture} shows a typical capture of a Nexus 6 rear camera in low light. This image was acquired for ISO 3000 and a 32 ms exposure time. The only image processing performed on this image was bi-linear demosaicking. The severe levels of noise present in the image demonstrate that low and medium light conditions represent a major challenge for imaging and computer vision systems. Note that particularly inexpensive low-end sensors will exhibit drastically worse performance compared to higher end smartphone camera modules.

In addition, we calibrated the optical aberrations $\psf$ from Eq.~\ref{eq:imaging_PSF} using a Bernoulli noise chart with checkerboard
features, following Mosleh et al. \shortcite{Mosleh_2015_CVPR} for spatially-varying PSF calibration.
The right plots in Fig.~\ref{fig:calibration} show the PSF $\psf$ for entire field-of-view of a Nexus 5 rear phone camera optic.
An in-depth description of our calibration procedure is provided in the Supplemental Material.
\changed{Alternative approaches to learned data generation for 
image reconstruction methods have been proposed in \cite{Brooks:2019,Jaroensri:2019}.}


\section{End-to-end Framework}
\label{sec:math}
In this section, we describe the proposed architecture for joint denoising, demosaicking,
(deblurring,) and classification. We evaluate the joint architecture in Sec.~\ref{sec:assessment}, as well as ablated models where only the low-level or high-level pipeline is trained or a conventional ISP pipeline is used. We assess the performance of the proposed model both on the simulated data and on \emph{captured RAW images}, to show that our simulated results carry over to real data.

\begin{figure*}[t]
	\centering
		\includegraphics[width=\textwidth]{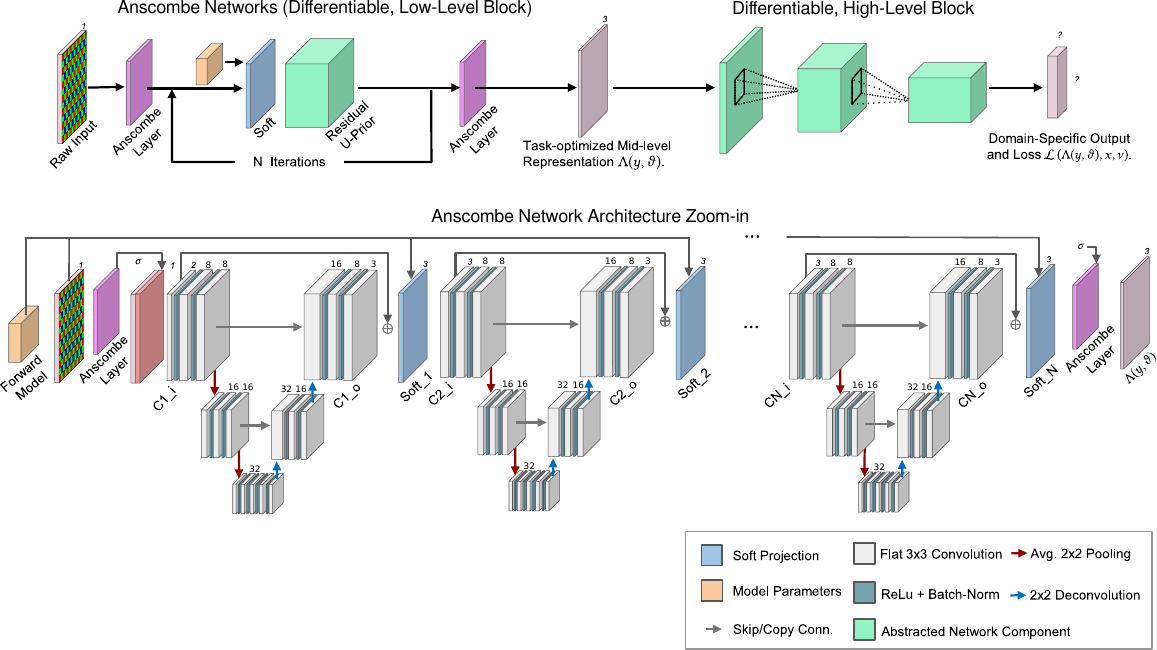}
		\caption{The proposed end-to-end architecture (top) for joint denoising, demosaicking, (deblurring) and
      classification combines a novel low-level Anscombe network block and a high-level task-specific network component in a
      single stack that takes in RAW CFA sensor data and outputs image labels.
      The Anscombe network component (zoom-in on the bottom) exploits knowledge of the calibrated image
      formation model and a learned proximal operator in a recurrent manner.
      The high-level model takes the output of the Anscombe network unit (either a feature tensor or an image) 
      and feeds it into a standard classification network trunk. This proximal operator in the Anscombe network 
      is a recurrent residual U-Net model with dense skip connections across all operator ``iterations''. 
      A partly unrolled network is show at the bottom.} 
		\label{fig:full-arch}
\end{figure*}

The architecture proposed in this work is illustrated in Fig.~\ref{fig:full-arch}. It combines jointly learned low-level and high-level processing units,
taking RAW sensor CFA data as input and outputting image labels. We propose a \emph{single differentiable model that generalizes across cameras and light levels}. This allows our model to abstract away the details of the camera for downstream applications, while being flexible and applicable to novel camera architectures.

We base the low-level block, which we dub Anscombe network unit, on an optimization algorithm $\Lambda$ that solves the problem of reconstructing an uncorrupted latent mid-level representation from noisy, single-channel, spatially-subsampled RAW measurements. In contrast to standard CNN models, the Anscombe layers in this model make the approach light-level independent and the unrolled optimization model achieves generalization across camera models (without retraining). We express the joint reconstruction and perception problem as a bilevel optimization problem
\begin{equation}\label{eq:bilevel_objective}
\begin{aligned}
&\;\minimize{\vartheta,\nu} \higherloss\left(\Lambda(y,\vartheta), x, \nu \right) \\
\textrm{s.t.} &\;\; \Lambda(y,\vartheta) = \; \argmin{x} \lowerloss\left(x, y,\vartheta \right),
\end{aligned}
\end{equation}
where $\Lambda$ minimizes here a lower-level objective $\lowerloss$. The output layer of this lower-level unit is an multi-channel mid-level representation $\Lambda(y,\vartheta)$, 
which is input into the higher-level model component and associated classification loss $\higherloss$. Here the model parameters $\nu$ of the higher-level model are absorbed in $\higherloss$ as a third argument.

For the nested objective $\lowerloss$, we follow a Bayesian approach as architecture backbone as it estimates a latent three-channel image $x$ exploiting both the probabilistic image formation model and allows for priors expressed in a principled fashion. \final{The Bayesian model assumes that $x$ is drawn from a prior distribution 
$\Omega(\vartheta)$, parameterized by $\vartheta$}.
We solve the Bayesian inference problem by unrolling an iterative optimization algorithm, only parameterizing the image prior with unknown, learned parameters, and truncating the iterations yielding the operator $\Lambda$.


%

Any differentiable higher-level image analysis method can be used in the proposed stack. In the following, we use the MobileNet-v1 classification network~\cite{mobilenet} as a higher-level network (which is replaced by a perceptual image loss when specializing the model for imaging for human vision~\ref{sec:results}). The higher-level classification loss $\higherloss$ is the standard cross-entropy classification loss. We chose the \revised{MobileNet} model family, since it is computationally efficient, running on modern smart-phone platforms in real-time, and while achieving competitive classification performance~\cite{mobilenet}. As the model is small, it can also be trained from scratch without data-center-scale training resources. Note that the proposed architecture can be adapted to other high-level computer vision tasks such as segmentation, object detection, and tracking, by replacing the classification network with another network for the given task. This also includes no high-level model, which then allows for the method to act as a learned ISP optimized for human viewing with adequate loss $\higherloss$, which we demonstrate in Sec.~\ref{sec:imaging_results}.

\subsection{Anscombe Networks}
The proposed low-level image processing unit, Anscombe networks, performs image 
reconstruction as a statistical estimation problem, which estimates a 
feature-preserving mid-level image from corrupted observations. 
We adopt a Bayesian approach and derive the proposed Anscombe network 
model as a maximum-a-posteriori (MAP) estimation method. 
As part of this model we introduce novel Anscombe network 
layers in this section, which allow for an efficient, compact, 
and transferable model that hence behaves like an 
ISP but is differentiable. 
\changed{Central to the design of our Anscombe networks are 
our variance-stabilizing Anscombe transform layers. \revised{Anscombe layers map Poisson-Gaussian distributed 
measurements $y$, to IID Gaussian 
noise~\cite{FM:13} with variance $\sigma=1$. 
Recall that the input to our Anscombe network, $y$, 
is the result of the 
camera image formation model defined in Eq.~\eqref{eq:imaging_PSF} (see calibration described in Section~\ref{sec:calibration}).
}
We show in Section~\ref{sec:results}
that Anscombe networks improve the accuracy of classifiers, trained 
with RAW data, that use conventional layers with capacity 
similar to our Anscombe networks.} 

In the Bayesian model, an unknown latent image $x$ is drawn from a prior distribution $\Omega(\vartheta)$ with parameters $\vartheta$. The linear transform $\sys$ from Eq.~\eqref{eq:imaging_PSF}, modeling all optical processes, transforms $x$ to the incident signal on the sensor, which is then measured by this sensor as an image $y$ drawn from a noise distribution $\omega(\sys x)$.
Recalling the image formation model from Eq.~\eqref{eq:imaging_PSF}, the transform $\sys$ models both the convolution with the kernel $\psf$ and subsampling on the CFA, and $\omega$ represents the calibrated Poisson-Gaussian noise.

Then the posterior probability of an unknown image $x$ yielding an observation $y$ is
\vspace{-3pt}
\BEQ
P(x|y;\vartheta) = \frac{P(y|\sys x)P(x;\vartheta)}{\int_{x}P(y|\sys x)P(x;\vartheta)}
\EEQ
with $P(y | \sys x)$ being the probability of sampling $y$ from $\omega(\sys x)$ and $P(x;\vartheta)$ be the prior probability of sampling $x$ from $\Omega(\vartheta)$. Because the posterior is proportional to $P(y|\sys x)P(x;\vartheta)$ the MAP estimate of $x$ is then given by
\vspace{-2pt}
\BEQ
x = \Argmax_x P(y|\sys x)P(x;\vartheta),
\EEQ
\vspace{-5pt}
or equivalently
\vspace{-2pt}
\BEQ\label{p-bayes}
  x = \Argmin_x \underbrace{f(y, \sys x) + r(x, \vartheta)}_{\lowerloss\left(x, y,\vartheta \right)},
\EEQ
where the data term $f(y, \sys x) = -\log P(y | \sys x)$ and prior $r(x,\vartheta) = -\log P(x;\vartheta)$ are negative log-likelihoods. These two terms define the lower-level objective $\lowerloss\left(x, y,\vartheta \right)$ from Eq.~\eqref{eq:bilevel_objective}.

\paragraph{Implicit Unrolled Proximal Optimization}

The low-level unit $\Lambda$ minimizes the loss $\lowerloss$ by solving Eq.~\eqref{p-bayes}. A large variety of algorithms have been developed for solving problem~(\ref{p-bayes}) efficiently for different convex data terms and priors, \eg, FISTA~\cite{doi:10.1137/080716542}, Chambolle-Pock~\cite{chambolle2011first}, ADMM~\cite{glowinski1975approximation}). The majority of these algorithms are iterative methods, in which a mapping $\Gamma(x^k, \sys, y, \vartheta) \to x^{k+1}$ is applied repeatedly to generate a series of iterates that converge to a solution $x^\star$, starting with an initial point $x^0$.

While an algorithm implementation can only be derived for explicitly given $f$ and $r$, we can define the algorithm itself with implicitly defined objectives. Suppose $f$ and $r$ are convex in $x$, and $r$ is differentiable. Then, we can solve Eq.~\eqref{p-bayes} with the proximal gradient method\final{~\cite{TB:09, doi:10.1137/080716542,parikh2013proximal,odp:17}}, which consists of the following updates
\begin{align}
x^{k+\sfrac{1}{2}} &= x^k - \alpha_k \nabla_x r(x^{k}, \vartheta) \label{prior-step}\\
x^{k+1} &= \Argmin_{x} f(y, \sys x) + \frac{1}{2\alpha_k}\|x - x^{k+\sfrac{1}{2}}\|^2_2,\label{data-step}
\end{align}
where $\alpha_k > 0$ is a step length. Each update consists of a prior step~\eqref{prior-step} and a data step \eqref{data-step}. The data step \eqref{data-step} is known as the proximal operator of $f$, that is
\BEQ\label{eq-prox}
\prox{\frac{\lambda}{\beta}f(y, \sys \cdot)}(x) = \Argmin_z \frac{\lambda}{\beta} f(y, \sys z) + \frac{1}{2}\|x - z\|^2.
\EEQ
Please see \cite{parikh2013proximal} for a detailed review of proximal operators and corresponding proximal optimization algorithms. One central idea that we rely on in this work is that we can also implicitly define steps of this algorithm. In particular, we propose to learn the prior mapping without explicitly defining the objective $r$, the space of all representations interpretable by the higher-level block, but rather parameterize the projection operator $\textrm{CNN}(\var,\theta^k) = \var - \alpha \nabla_x r(\var, \vartheta)$ with $\vartheta$ and $\alpha$ being learned implicitly.

Solving Eq.~\eqref{p-bayes} using an iterative optimization algorithm of the reader's choice would lead to an algorithm with a data-dependent termination criterion and no obvious method to learn unknown algorithm parameters since computing the derivatives of the output with respect to the algorithm parameters $\vartheta$ is value-dependent. An alternative approach is to execute a pre-determined number of iterations $N$, in other words unrolling the optimization algorithm. This approach is motivated by the fact that for many imaging applications very high accuracy, \eg, convergence below tolerance of $10^{-6}$ for every local pixel state, is not needed in practice, as opposed to optimization problems in, for instance, control. Instead, many applications are runtime-constrained, and truncation allows for a fixed runtime. Fixing the number of iterations allows us to view the iterative method as an explicit function $\Gamma^N(\cdot, \sys, y, \vartheta) \to x^N$ of the initial point $x^0$. Parameters such as $\vartheta$ may be fixed across all iterations or vary by iteration. The unrolled iterative algorithm can be interpreted as a deep network, and, if each iteration of the unrolled optimization is differentiable, the gradient of $\vartheta$ and other parameters with respect to a loss function on $x^N$ can be computed efficiently through backpropagation. The proposed network recipe is given in Algorithm~\ref{alg-prox-grad}. Note that we allow all parameters to differ across layers. The model is differentiable in its output with respect to each layer's free parameters.

\paragraph{Anscombe Layers}

\begin{algorithm}[t!]
	\caption{Anscombe networks: Variance-stabilized implicit proximal gradient network.}\label{alg-prox-grad}
	\begin{algorithmic}[1]
		\State $\tilde{y}, \sigma \gets \ansc(y)$
		\State{Recurrent vars: $x^0 = A^Ty$, $\alpha_k = C_0 C^{-k}$, $C_0> 0$, $C > 0$}
		\For{$k=0$ to $N-1$}
		\State $x^{k+1/2} \gets \textrm{CNN}(x^k,\vartheta^k)$.
		\State $x^{k+1} \gets \Argmin_x \alpha_{k} f(Ax,\tilde{y}) +  \frac{1}{2}\|x -  x^k - x^{k+1/2}\|^2_2$.
		\EndFor
		\State $\tilde{x} \gets \ansc^{-1}(x^N, \sigma)$
	\end{algorithmic}
\end{algorithm}
The network generated by Algorithm~\ref{alg-prox-grad} is an implicit unrolled proximal gradient network. However, rather than working directly on the measurements $y$, which are Poisson-Gaussian distributed according to Eq.~\eqref{eq:imaging_PSF}, we embed the unrolled architecture in variance-stabilizing Anscombe transform layers, converting the Poisson-Gaussian noise into IID Gaussian noise \cite{FM:13} with variance $\sigma=1$. This has the benefit that the data step in line 5 of Alg.~\ref{alg-prox-grad} becomes a simple quadratic term, and image features at all intensity levels are affected by the same noise degradations, effectively regularizing the model to perform robustly independent of the light level.

Specifically, we apply the generalized Anscombe transform~\cite{FM:13} as a first layer, denoted by the operator $\mathcal{A}$, to the measured single channel RAW observation $y$, The transform and \revised{its} unbiased inexact inverse are defined as
\begin{align}
  \ansc & : x \mapsto 2 \sqrt{x + \tfrac{3}{8}}, \label{eq:ansc} \\
	\ansc^{-1} & :x \mapsto \frac{1}{4} x^2 - \frac{1}{8} - \sigma^2. \label{eq:invansc}
\end{align}
However, RAW data input to this transform, without modifications, results in peak signals that are not consistent across training examples. Hence, the gradient components of the unrolled proximal gradient method are not normalized with respect to light level, leading to poor network performance. To avoid this behavior, we max-normalize the output of the forward Anscombe transform with the multiplicative factor $s_\ansc=1/\max(\ansc{x})$. While this normalizes the value range to the interval $[0,1]$, the unit-variance Gaussian noise distributed $\ansc{x}$ becomes Gaussian-distributed with variance $\sigma = s_\ansc^2$. As this parameter is known, we provide it to the network as a separate channel, which is illustrated in Fig.~\ref{fig:full-arch}. The output of the unrolled proximal gradient network component followed by the inverse generalized Anscombe transform $\mathcal{A}^{-1}$, which inverts the shift and scaling, then applies the inverse transform. The noise parameters are known from the ISO and the precalibrated noise curves from Sec.~\ref{sec:calibration}.

\paragraph{Soft Projection Layers}

\revised{The data step in Alg.~\ref{alg-prox-grad} (line 5)} is the ``soft projection'' operator
$\Pi(\cdot,\gamma,\sys,y)$ given by
\[\label{eq:soft-proj}
  \Pi(v,\gamma,\sys,y) = \Argmin_z \frac{1}{2}\|y - \sys z\|^2_2 + \frac{\gamma}{2}\|v - z\|^2_2.
\]
Recalling Eq.~\eqref{eq-prox}, $\Pi(\cdot,\gamma,A,y)$ is the proximal operator of the function $f$. With the Anscombe layers present, this function, i.e. the negative log-likelihood $-\log P(y | \sys x)$ from Eq.~\eqref{p-bayes}, becomes a simple quadratic now, that is
\[\label{eq:data_term}
f(y, \sys x) = \frac{1}{2}\|y - \sys x\|^2_2.
\]
Hence, the operator $\Pi$ can be computed efficiently as an unconstrained quadratic optimization problem. In the case of joint demosaicking and denoising, the operator $\sys=\cfa$ and $\Pi$ becomes
\[\label{eq:soft-proj-subsampling}
\Pi(v,\gamma,\cfa,y) =  \frac{\cfa^T \ansc{y} + \gamma z}{\cfa \ones + \gamma},
\]
where division is elementwise. The soft projection parameter $\gamma > 0$ trades off closeness to the input $v$ with fidelity to the measurements $y$ (\ie, ensuring $y \approx Ax$). We dub this operator ``soft projection'' because in the limit $\gamma \to 0$, $\Pi(v,\gamma,A,y)$ is the Euclidean projection of $v$ onto the linear system $y = \sys x$. Note that $\gamma$ may be learned or fixed. 

The soft projection data step is inspired by analysis in the Supplemental Material. We found that substantially improved generalization over naive residual CNN models could be achieved due to applying soft projection in the proposed unrolled architecture, in particular for tasks where the imaging operator $\sys$ varies across camera or example (for instance, different CFA patterns, optical systems, or images that are blurred with different blur kernels). Intuitively, soft projection decouples the (approximate) inversion of the physical image
operator $\sys$ from the prior step. Thus, the model does not have to re-learn the (approximate) inversion of $\sys$ depending on sensor, optical parameters, or capture settings, and need instead only learn prior parameters and algorithm hyper-parameters.

\subsection{Residual U-Net Prior Parametrization}

The purpose of the cascade of prior network units in our architecture is to map estimates of the unknown midlevel representation $x$ onto a nearby point in the manifold of representations that are interpretable by the higher-level network, or when optimizing for human viewing (i.e. the space of perceivable natural images). The prior steps from Algorithm~\ref{alg-prox-grad} must therefore be flexible enough to learn the complex statistics of natural images but also project on a subset according to the higher-level loss $\higherloss$. 

We use a CNN as learnable prior architecture, as CNNs are established architectures for feature-encoding in the image domain and are thus a natural choice for learning the mapping onto the subset manifolds of natural images. Specifically, we propose a deep residual U-Net~\cite{unetRonneberger} variant with three levels (see Figure~\ref{fig:full-arch}), $3\times3$ convolution kernels, ReLU nonlinearities, downsampling with $2\times2$ average pooling, upsampling by $2\times2$ deconvolution layers (transpose convolution), and batch normalization in the intermediate layers to ease training~\cite{ioffe2015batch}. The number of channels in the first U-Net level increases by a factor of 2. The channels are doubled in each of the three levels of the U-Net. The U-Net prior at iteration $k$ in the unrolled stack takes as input the output of the soft projection step $k-1$ concatenated with the Anscombe noise parameter $\sigma$ as a separate channel. In order to handle the RAW color-filter array data, the very first layer in the U-Net prior at iteration $0$ uses a stride 2 convolution in the very first convolutional layer. Further information on the U-Net parametrization can be found in the supplement.

We note that the U-Net priors are trained end-to-end as part of the complete architecture in Fig.~\ref{fig:full-arch} and a different prior is trained for each iteration of the unrolled optimization stack. This allows each prior step to specialize in removing \emph{correlated noise}, i.e.~reconstruction artifacts, introduced by the preceding data step, such as inpainting artifacts aligned with the CFA or inverse filtering ringing artifacts.

\section{Evaluation}
\label{sec:assessment}
\begin{table*}[ht!]
  \centering
	\ra{1.3}
	\resizebox{\textwidth}{!}{%
	\changed{
	\begin{tabular}{@{}rcccccccccccccc@{}}\toprule
		& \multicolumn{2}{c}{3 lux} & \phantom{a}& \multicolumn{2}{c}{6 lux} &
		\phantom{a} & \multicolumn{2}{c}{2 to 20 lux} & \phantom{a} &
                      \multicolumn{2}{c}{2 to 200 lux} & \phantom{a} &
                    \multicolumn{2}{c}{Size and Complexity}\\
		\cmidrule{2-3} \cmidrule{5-6} \cmidrule{8-9} \cmidrule{11-12} \cmidrule{14-15}
		& Top-1 & Top-5 && Top-1 & Top-5 && Top-1 & Top-5 && Top-1 & Top-5 && \# Params. (M) & FLOPs (M) \\ \midrule \midrule
		From Scratch MobileNet-v1 (RAW input w/o ISP) & 23.53\% & 44.13\%  && 32.65\%  & 55.94\% && 35.16\% & 58.14\% && \revised{\underline{42.65}\%} & \revised{\underline{66.13}\%} && 4.23 &  181 \\
		From Scratch deeper MobileNet-v2 (RAW input w/o ISP) & \revised{\underline{27.87}\%} & \revised{\underline{50.82}\%}  && \revised{\underline{36.80}\%}  & \revised{\underline{61.31}\%} && \revised{\underline{36.32}\%} & \revised{\underline{59.40}\%} && 38.56\% & 61.58\% && 6.90 & 320 \\
		Movidius Myriad 2 ISP + Pretrained & 0.22\% &1.10\% && 1.69\% & 5.39\% && 9.12\% &
                                 18.63\% && 17.78\%  & 32.11\% && 4.23 & 181 \\
		Darktable ISP + Pretrained & 0.22\% &0.46\% && 0.46\% & 1.52\% && 7.12\% & 15.28\% &&
                                                                                 18.20\% & 32.04\% && 4.23 & 181 \\
		Movidius Myriad 2 ISP + Finetuning & 23.52\% & 44.69\% && 36.11\% & 60.27\% && 17.31 \% &  34.98 \%
                                                                && 21.75\% & 40.93 \% && 4.23 & 181 \\
	 	Darktable ISP + Finetuning & 20.02\% & 39.56\% && 34.73\% & 58.55\% && 16.45 \% &  33.13\%  && 23.47\% & 43.77\% && 4.23 & 181 \\		 		
		U-Net~\cite{chen2018learning} + Finetuning & 29.89 \% & 52.62 \%  && 36.20\%  & 60.23\% && 14.38\% & 30.44\% && 20.23\% & 39.35\% && 11.99 & 537 \\
		U-Net~\cite{chen2018learning} + percep. loss + Finetuning & 9.52\% & 20.84\%  && 25.74\% & 45.81\% && 19.09\% & 36.26\% && 26.17\% & 46.78\% && 11.99 & 537 \\ \midrule \midrule
		Proposed Joint Architecture (\revised{MobileNet}-v1 Head) & \textbf{30.50\%} & \textbf{53.28\%} && \textbf{43.63\%} & \textbf{67.73\%} && \textbf{40.87\%} & \textbf{64.01\%} && \textbf{48.46\%} & \textbf{71.44\%} && 4.28 & 282 \\
		\bottomrule
	\end{tabular}
	}
	}
  \caption{Classification results on simulated data. We compare the proposed joint architecture to classifiers that ingest (and are trained on) pre-processed images output by conventional ISPs, including Darktable and the Movidius Myriad 2 ISP, \changed{and learnable deep ISPs~\cite{chen2018learning}}. 
  Off-the-shelf MobileNet-v1 classifiers pretrained on Imagenet perform poorly on ISP-preprocessed 
  data (Movidius Myriad 2 ISP + Pretrained and Darktable ISP + Pretrained). Fine-tuning these classifiers on the ISP-processed data (Movidius Myriad 2 ISP + Finetuning, Movidius Myriad 2 ISP + Finetuning, \changed{and U-Net~\cite{chen2018learning} deep ISP models}) results in substantially improved performance. \changed{While the parameters of the conventional ISPs have been expert-tuned, we train the deep U-Net ISP from~\cite{chen2018learning} on the clear/noisy training corpus, and we also report results when adding an perceptual loss~\cite{johnson2016perceptual} (+ percep. loss). However, none of the fine-tuned models, trained on top of traditional or learnable ISPs, does outperform 
  networks that \emph{do not employ an ISP at all} across all settings, as evidenced by results of a MobileNet-v1 on unprocessed RAW data (From Scratch MobileNet-v1).} 
   Only the proposed joint architecture with a learned Anscombe network 
  outperforms both, traditional pipelines, as well as from-scratch-trained models in both Top-1 and Top-5 classification accuracy across illumination conditions. \changed{The proposed approach even outperforms from-scratch-trained MobileNet-v2 models that are \emph{deeper networks with larger network capacity} compared to our architecture. Note that {all other models} compared in this table use the MobileNet-v1 architecture as classifier heads.} \revised{We highlight the best and second best models using bold and underlined text, respectively.}}
	\label{tab:benchmark}
	\vspace{-12pt}
\end{table*}

Next, we describe the evaluation of our proposed methods.
First, we evaluate our joint imaging and perception model on 
classification of low-light RAW images. Specifically, we captured 
a new dataset over a range of low-light levels and also built 
a synthetic low-light dataset based on ImageNet. We include ablated studies that 
show the importance of our proposed low-level 
Anscombe network to improve the high-level network accuracy. Second, 
we evaluate our low-level model for image reconstruction in low-light 
for human viewing (imaging without considering a high-level 
task, i.e. classification). For this evaluation, 
we use a recent publicly available dataset~\cite{chen2018learning}, 
which consists of short and low exposure images. Third, we demonstrate 
a real time mobile prototype implemented using the Android Camera2 API 
and a remote Tensorflow model server. \final{The next sections} describe 
the experimental setup and results found over these evaluations.


\subsection{Evaluation of Low-light Imaging and Perception -- Synthetic Data}
\label{sec:results}

We trained instances of the proposed joint architecture for four challenging scenarios: 
3 lux, 6 lux, the range 2 to 20 lux, and the range 2 to 200 lux. \changed{While the first two settings allow us to analyze the models 
in specific low-light conditions, the scenarios with ranges of illuminance allow us to evaluate the generalization of the models over a variety of different light levels.} Specifically, we trained and evaluated the models on \revised{a noisy version of ImageNet, constructed
using the image 
formation model from Sec.~\ref{sec:image_formation}, calibrated for the Nexus 5 rear camera 
for a given light-level (or a light-level sampled randomly from a range)}.
To evaluate the effect of noise separately from optical aberrations, we ignore aberrations 
in the following (see Supplemental Material).
The results reported next correspond to the ImageNet validation 
set of 50,000 images \cite{deng2009imagenet}\revised{, which consists of 
1000 object classes, and 50 samples per class.}

\changed{
To evaluate over many different noise settings and to be able to train deep nets completely from scratch (Table~\ref{tab:benchmark}), 
we opt to use the computational efficient MobileNet classification network 
in all the following experiments. \revised{We refer the reader to the Supplementary document 
for results using the much larger Inception-v4 classifier on a smaller subset 
of the evaluations taking one month of training time.}
We compare the proposed joint architecture (Joint Anscombe Network and MobileNet-v1) 
to the following baselines:
}

\begin{itemize}
 \item \changed{The conventional approach of combining a high-quality ISP, 
optimized for human viewing, with an existing 
pretrained MobileNet-v1 classifier. 
}

\item \changed{Using a trainable state-of-the-art ISP~\cite{chen2018learning}, 
fine-tuned for image quality in each noise scenario, and 
a MobileNet-v1 classifier finetuned on the learned 
ISP output.}

\item \changed{A MobileNet-v1 classifier directly trained from scratch on RAW noisy data.}

\item \changed{As a deeper version of our classifier with higher model capacity, 
we train the \revised{MobileNet}-v2 (1.4) classifier~\cite{Sandler:2018} from scratch, with 50\% more parameters and 
about 40 million more FLOPS.}


\end{itemize}

\changed{We train all the evaluated models until convergence with large iteration buffer.
Table~\ref{tab:benchmark} summarizes the results for the described low-light scenarios. 
We next describe our findings from this evaluation.}

%


\changed{
\paragraph{Combining high-quality ISPs
with pretrained high-level network fails in low-light.}
In this experiment, we use the hardware ISP of a Movidius Myriad 2 
evaluation board, and the high-quality open-source ISP 
Darktable~\shortcite{darktable} both engineered and optimized for visual image quality. We note that the Darktable 
uses a non-local means block-matching denoiser (NLM)~\cite{buades2005non} 
that is prohibitively costly to implement in hardware. The parameters of the darktable RAW 
developing tool and the Movidius Myriad 2 were hand-tuned by a human expert to maximize perceptual quality.
The results in Table~\ref{tab:benchmark}, rows third and fourth, 
validate that the conventional approach of combining a high-quality ISP, 
optimized for human viewing, with an existing pretrained high-level 
network fails in low-light scenarios. 
\emph{In all cases}, this approach performs weakly due to the 
severe noise present in the image data. 
This applies both to efficient hardware ISP architectures, 
such as the Movidius Myriad 2 ISP, as well as to high-performance 
photography RAW processing ISPs, such as Darktable. In fact, 
as detailed below, processing RAW measurements with conventional image 
processing units, tuned for perceptual quality, can decrease classification 
performance, compared to almost unprocessed bi-linearly interpolated color images. 
These findings also apply to image degradations introduced by optical aberrations. 
We refer the reader to the supplement for a study on the effect of optical aberrations.
}

\paragraph{\changed{Finetuning a classifier on top of ISP-preprocessed images  
does not do better than a model trained directly on RAW noisy data.}}
\changed{Traditional ISP pipelines achieve acceptable performance only} when 
the networks are fine-tuned, i.e. specialized, to the degraded low-light 
imaging data output by the respective ISP 
\changed{(fifth and sixth rows of Table~\ref{tab:benchmark}). 
However, the performance of these specialized networks applied on the output of existing ISPs is 
exceeded by simply training a network from scratch for the given imaging 
condition but without an ISP at all, only using bilinearly 
demosaicked color images \changed{(first row of Table~\ref{tab:benchmark})}.} 
The classifier trained without ISP preprocessing obtained higher Top-1 
accuracy on three out of the four noise settings. 
\changed{Overall, processing images with conventional ISP pipelines, that are designed 
and tuned for human viewing, at best marginally increased classification 
accuracy for models specialized to individual light levels and in many cases substantially decreased performance. 
This is especially apparent for varying 
low-light conditions (columns 2-to-20 lux and 2-to-200 lux), where 
classifiers finetuned on ISP outputs obtain only half of the Top-1 
accuracy of classifiers trained from scratch.}
On a first glance, this result may argue for completely removing 
traditional ISP pipelines and simply train standard CNN classifiers 
with as little traditional preprocessing as possible. 

\paragraph{\changed{Anscombe Networks outperform classifiers trained from scratch on RAW data (even with larger model capacity).}}
\changed{We compare the proposed method to \revised{MobileNet-v1} trained directly on RAW data and its deeper larger variant, MobileNet-v2; see first, second and last 
rows of Table~\ref{tab:benchmark}. The \revised{MobileNet-v2} variant 
introduces inverted residual blocks, in which shortcut connections 
are introduced between bottleneck layers, and improves efficiency and accuracy relative to 
MobileNet-v1. Specifically, we use MobileNet-v2 (1.4)~\cite{Sandler:2018}, 
that has larger capacity (1.4 width multiplier) than the standard version.
We observe that this deeper model improves results in almost all illumination conditions compared to \revised{MobileNet-v1}. For the larger 2-to-200~lux illumination range, we do observe worse performance which we attribute to the larger architecture being slightly more prone to overfitting as a result of memorization.} 
\changed{
We note that our joint network obtains higher Top-1 and Top-5 accuracy compared to both models across all 
noise settings. Although MobileNet-v2 has a substantial higher number of parameters, 
this larger capacity does not translate into an improvement over our joint models.
Finally, note that for both MobileNet networks, 
there is not explicit modeling of an 
intermediate image. These results validate that the 
improvement obtained by our 
joint architecture does not come from a larger capacity compared to the MobileNet-v1 version, 
but from the design of our Anscombe network.
As such, we demonstrate that Anscombe Networks are highly 
effective at recovering an intermediate image representation that are tailored to the downstream task across different noise scenarios.
}

\paragraph{\changed{Anscombe Networks outperform classifiers trained on top of
learnable deep ISP outputs.}}
\changed{
As a further comparison, we finetune a classifier network also on the outputs of existing learnable deep ISPs. Specfically, we train the deep ISP network from Chen et al.~\shortcite{chen2018learning} for image reconstruction with the loss settings proposed by the authors. We then finetune a MobileNet-v1 network on a corpus of denoised images. The results of this experiment are shown in the seventh row of 
Table~\ref{tab:benchmark}, validating that our network also outperforms 
this approach for all noise scenarios. The margins are especially high for varying light levels (2-to-20 lux and 2-to-200 lux).}

\changed{In addition, we also provide results using a perceptual loss~\cite{johnson2016perceptual} in the first stage of the deep ISP training in addition to the $\ell_1$-loss proposed in~\cite{chen2018learning}. We manually optimized the weight ratios of both objective components.
While this perceptual loss adds robustness across light level ranges, the proposed model maintains a high margin over this baseline. These results emphasize the efficiency and effectiveness of Anscombe Networks, 
which have \emph{$2.5 \times$ and $2 \times$ fewer parameters and 
FLOPs}, respectively, compared to the learnable baselines.
}

\paragraph{\changed{Computational Complexity.}}
\changed{
The last two columns of Table~\ref{tab:benchmark} list the 
computational complexity of all models.
For the deep ISP~\cite{chen2018learning} with perceptual loss~\cite{johnson2016perceptual}, 
we do not consider the additional parameters of the pre-trained classifier used in the perceptual loss calculation, and for the Movidius and Darktable ISPs we only measure the MobileNet-v1 network compute cost, although modern ASIC ISPs require tremendous engineering efforts to be power efficient. For all models, we list complexity for RAW input images of 128x128 size. 
We note that the proposed joint architecture consists of 
\emph{only $0.1 \times$ additional parameters (Anscombe network)} relative to 
MobileNet-v1, and this represents $2.5 \times$ fewer parameters than the deep ISP from~\cite{chen2018learning}. Our joint architecture 
runs at 60fps.}

\paragraph{\changed{Robustness and Generalization.}}
\changed{The results in Table~\ref{tab:benchmark} validate} the effectiveness of our proposed joint 
architecture to recover relevant information from RAW data 
to \emph{improve accuracy for a high-level task}. We also emphasize 
our model's outstanding \emph{generalization across different 
light levels}. We can see in Table~\ref{tab:benchmark} that while the 
accuracy of the models that use state-of-the-art \changed{software, hardware, and learnable} 
ISPs drastically decrease over the 2-to-20 and 2-to-200 lux ranges, the accuracy of our proposed model 
remains stable. This again underlines the 
limitations of conventional models under more realistic scenarios, 
where light levels are highly variable, and the importance of building 
generalizable models. \changed{Our models also shines when comparing 
computational complexity, enabling robust real-time applications, as shown in Section~\ref{sec:prototype}.}

\begin{figure*}[htp]
	\centering
	\includegraphics[width=0.93\textwidth]{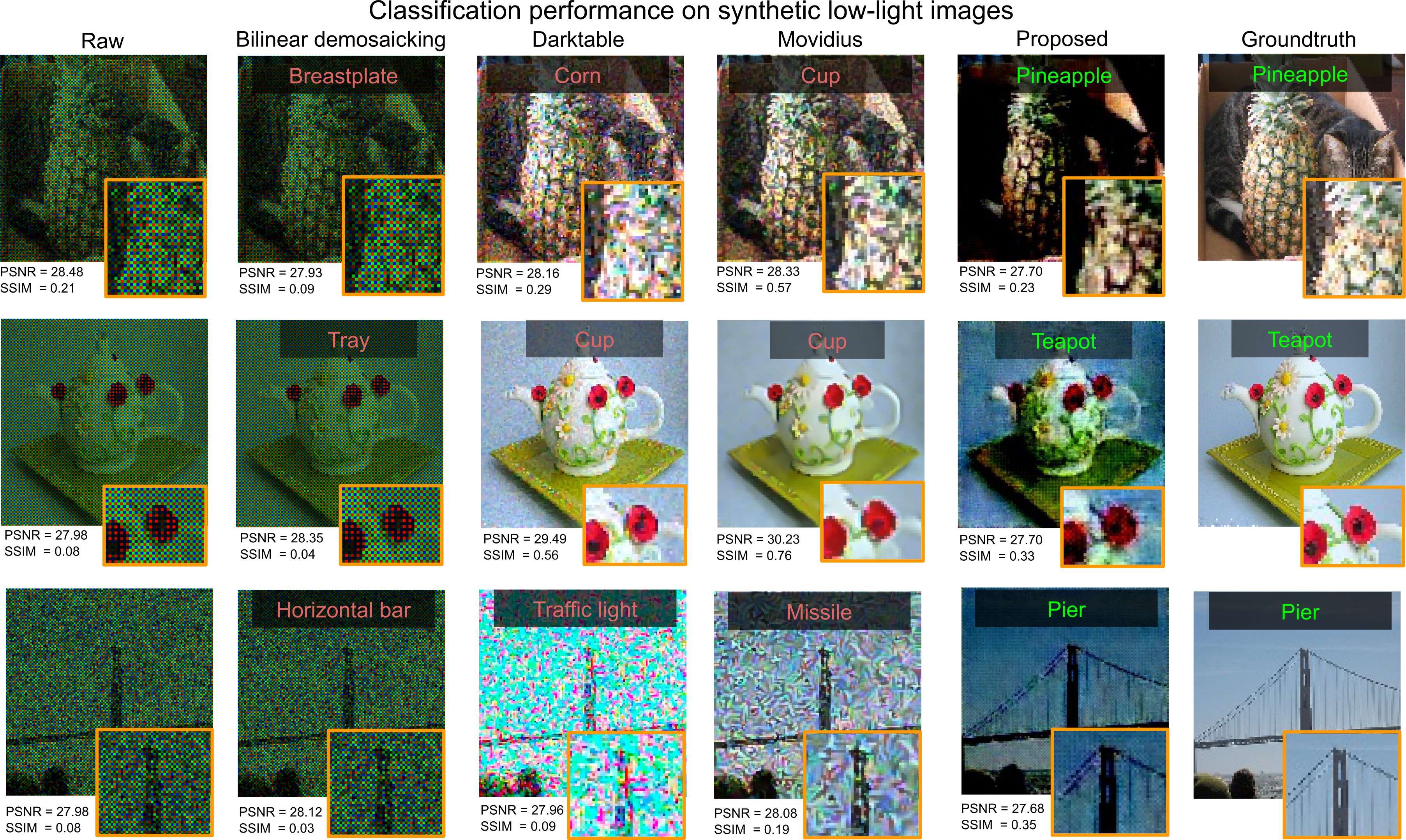} \\ \vspace{5mm}
	\includegraphics[width=0.93\textwidth]{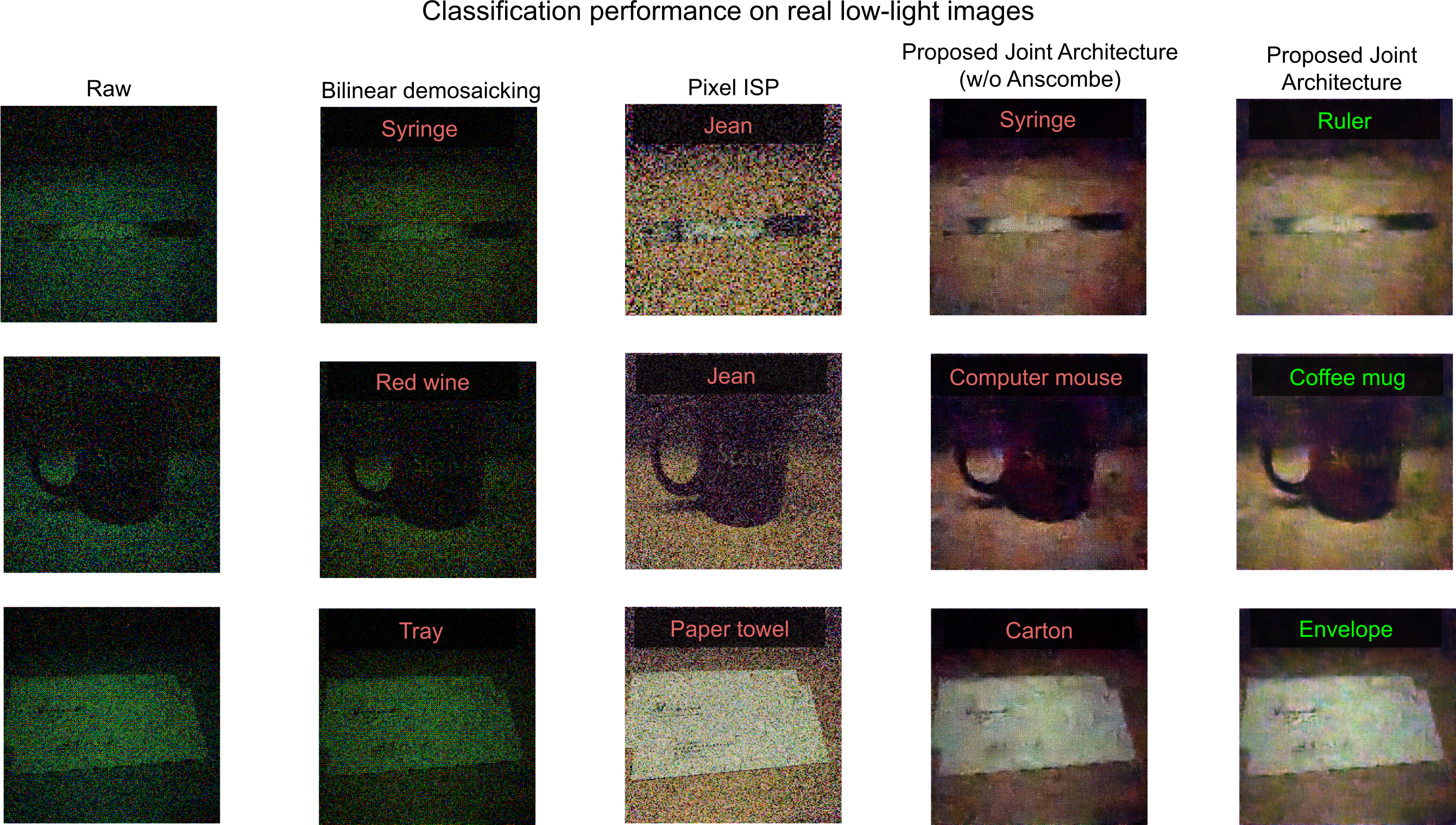}
	\caption{
	{Synthetic and real low-light RAW images, and their corresponding classification results after processing with 
	conventional ISPs (Darktable and Movidius), bilinear demosaicking, and Anscombe Networks.
	The proposed joint architecture scores lower in terms of PSNR and SSIM. However, 
	\changed{results suggest that}
	our proposed model does not only removes noise, but selectively amplifies structures of the target class, 
	which seems to benefit the overall classification accuracy of the model}.	
	%
		}
	\label{fig:denoiser-qual}
\end{figure*}

\vspace{-3pt}
\paragraph{Qualitative Interpretation.}
The results in Table~\ref{tab:benchmark} raise the question of why the jointly training Anscombe networks was so much more helpful to the classifier than
conventional algorithms.
The images in Fig.~\ref{fig:denoiser-qual} suggest an answer. 
Fig.~\ref{fig:denoiser-qual} shows a low-light image that was incorrectly classified by the pretrained MobileNet network
but correctly classified by the joint architecture\footnotemark.
The RAW input image and a bilinearly demosaicked image is shown, 
as well as the outputs of the conventional hardware and software ISPs, 
and the intermediate mid-level representation produced by the 
Anscombe network unit. The label assigned by the classifier 
is given in each instance, as well as the PSNR and SSIM 
relative to the original image.

The images output by conventional ISPs for human viewing
contain less noise than unprocessed RAW data.
Fine details of the target class are blurred out, however.
\changed{Comparing conventional and learned ISP outputs with Anscombe network's intermediates, 
we hypothesize that our joint Anscombe architecture 
tailors processing to the classification task by 
selectively boosting contrast around structures of the target class while 
removing noise in large smooth regions. This selective processing 
seems to be key to recover the target class structures independently of the noise or light level, which explains the robustness of our model across different light levels.}

\changed{As a result, by conventional metrics of restoration quality such as SSIM and PSNR,
the joint unit is, in fact, worse than conventional algorithms. These metrics do not distinguish between scene content necessary for a classification and background regions without task-specific information.
We can also see, though that it preserves and amplifies detail that is useful to the
classification network, the proposed Anscombe network does perform denoising and deblurring of the image.
The qualitative results suggest traditional reconstruction algorithms and metrics used to make
images visually pleasing to humans are not appropriate for high-level analytic tasks.}

\vspace{-3pt}
\subsection{Evaluation in Low-light Imaging and Perception -- Captured RAW Data}
We demonstrate generalization of the proposed models to real-world low-light images. 
Using a Google Pixel phone rear camera, we collected low-light image patches in 
the wild. Rather than adopting the lengthy process of extracting these patches 
from objects at various scales in arbitrary photographs, we acquire full-frame 
images that directly correspond to classes in the ImageNet dataset and create patches 
by subsampling. While not affecting per-pixel noise, this process enables us to 
eliminate the effect of blur in the capture, allowing us to make solid claims about 
the effect of noise in isolation. 
\revised{The same applies to demosaicking which typically only considers a small neighborhood 
of pixels}.
We collect a low-light dataset approximately 
corresponding to light levels between 1 lux and 200 lux. The dataset 
consists of 1103 images across 40 imagenet classes respectively 
%
\footnotetext{\changed{Please see Supplemental document for additional visualizations of the finetuned outputs of the 
deep ISP from~\cite{chen2018learning}.}}
\begin{table}[t!]
  \centering
	\ra{1.3}
	\resizebox{\columnwidth}{!}{%
	\changed{
	\begin{tabular}{rccccccccc}\toprule
		& Top-1 & Top-5 & \#Parameters & FLOPS \\ \midrule \midrule
		From Scratch MobileNet-v1 & 27.03\% & 52.45\% & 4.23 &  181 \\
		\revised{From Scratch MobileNet-v2} & \revised{26.92\%} & \revised{\underline{56.45}\%} & \revised{6.90} &  \revised{320} \\
		Pixel ISP~\footnotemark + Pretrained MobileNet-v1 & 1.4\% & 14.1\% & 4.23 &  181 \\		
		U-Net + Anscombe layers + MobileNet-v1 & \underline{28.80}\% & 55.20\% & 11.99 & 537  \\ \midrule \midrule
		Proposed Joint Architecture (no Anscombe layers) & 28.53\% & 54.25\% & 4.28 & 282  \\
		Proposed Joint Architecture  & \textbf{33.13\%} & \textbf{58.36\%} & 4.28 & 282  \\
		\bottomrule
	\end{tabular}
	}
	}
	\caption{Results on data captured in the wild with a Google Pixel phone rear
    camera for models trained on 2 to 200 lux. The exposure time was fixed at
    1/10000 and the ISO at 8000. Additional digital gain was applied to
    normalize brightness. \revised{The best and second best methods are highlighted in bold and underlined text, respectively.}\vspace{-18pt}}
	\label{tab:raw_benchmark}
\end{table}
\footnotetext{\changed{We do not count the parameters and FLOPS of the proprietary Pixel ISP here.}}
\noindent
\changed{Table~\ref{tab:raw_benchmark} lists results 
on the real-world dataset, including ablations of our proposed architecture.} 
\revised{The evaluated models correspond to the 2 to 200 lux models in Table~\ref{tab:benchmark}. These experiments evaluate the generalization 
performance of the respective models to real captured data.}
Absolute performance 
is worse than on the simulated datasets, which is 
likely due to a mismatch between how classes appear in ImageNet and how 
they appear in the wild. 
The relative margins are consistent with 
the simulated results.

\vspace{-8pt}
\changed{
\paragraph{Anscombe networks generalize well to real data}
Table~\ref{tab:raw_benchmark} confirms that the highest classification 
accuracy was achieved by the 
proposed joint model, with Top-1 and Top-5 accuracy up to 6\% higher 
than the fine-tuned models. 
The from-scratch tuned MobileNet-v1 \revised{and MobileNet-v2 models} outperform the pretrained 
MobileNet network on Pixel ISP substantially, by 10s of percent.
}

\vspace{-8pt}
\changed{
\paragraph{Ablations of the proximal operator and Anscombe layers.}
Table~\ref{tab:raw_benchmark} also includes results of the joint architecture 
\revised{without the proximal operator network (fourth row) and without the 
Anscombe transform (fifth row)}. For additional comparison, \revised{the network 
without the proximal operator uses} the larger U-Net architecture 
described by~\cite{chen2018learning} while keeping the Anscombe transform at the input and \revised{its} inverse at the output 
of this network. 
These experiments validate that the architecture that uses Anscombe
transform outperfoms the one that does not include this transform 
by around 5\% in both Top-1 and Top-5 accuracy. Also, 
replacing the proximal operator network with U-Net reduces the accuracy
of our proposed model by 4\% and 3\% in Top-1 and Top-5 accuracy, respectively.
This margin validates that the Anscombe network as a whole is key for the performance of the overall joint model including the high-level classification model.
}

\begin{figure*}[t!]
\vspace{-6pt}
	\centering
	\includegraphics[width=\linewidth]{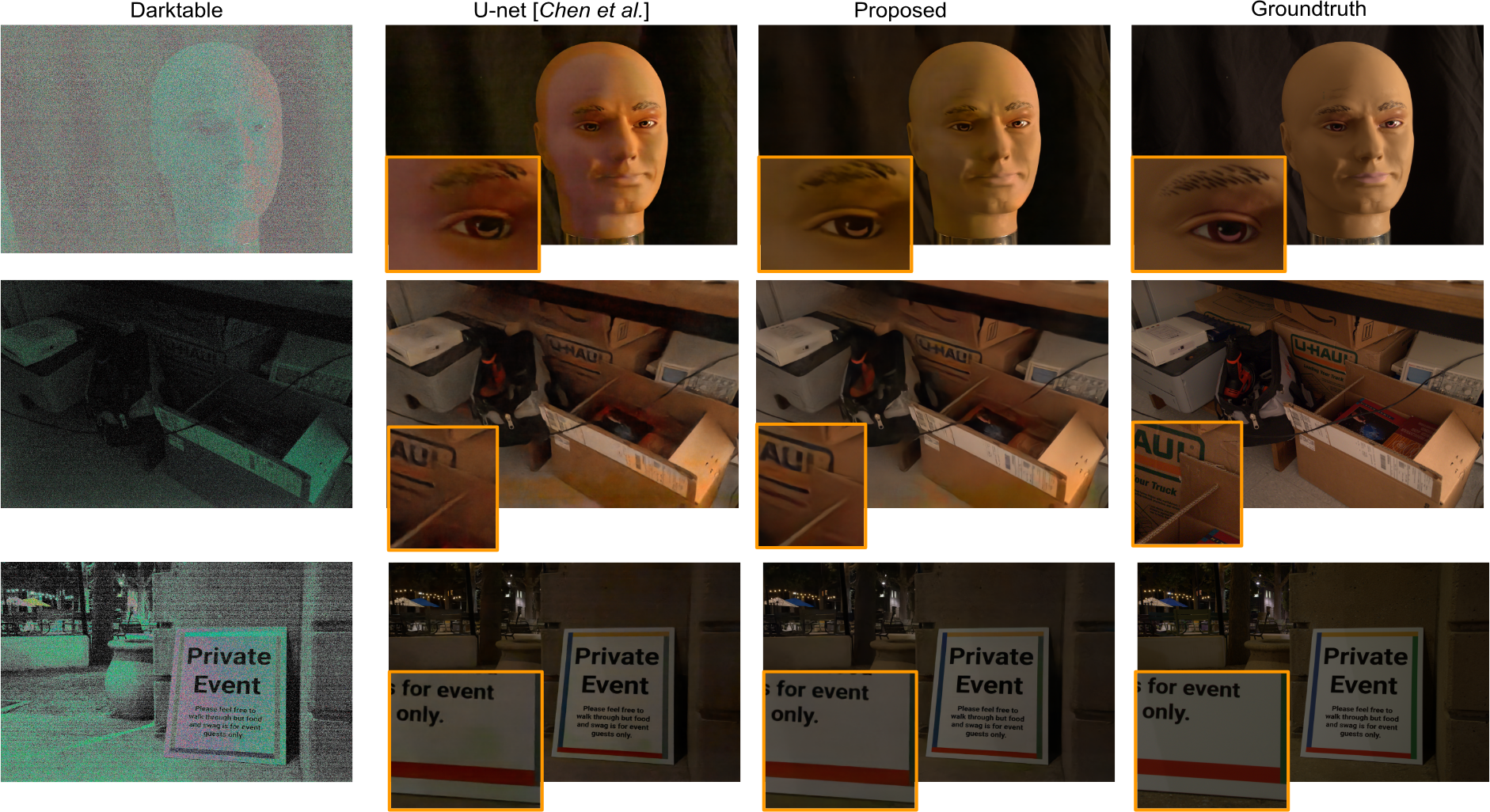}
	\caption{Anscombe Networks for Single-Image Low-Light Photography (w/o Classification). Qualitative low-light denoising results for human viewing using the traditional Darktable ISP, the U-Net model 
	proposed by~\cite{chen2018learning}, and the proposed Anscombe network. The proposed model and Chen et al.~\shortcite{chen2018learning} have been trained on the same low-light dataset 
	from the dataset proposed in~\cite{chen2018learning}. All methods use the same single RAW image as input. 	
		}
	\label{fig:RAW_image_processing}
\end{figure*}

\paragraph{Qualitative Results}
Fig.~\ref{fig:denoiser-qual} helps \changed{to} explain the improvement in classification 
accuracy of the proposed joint model as compared to conventional+fine-tuned and 
from-scratch baselines. We show image examples that each went through four different 
classification pipelines: one without any processing except for bilinear demosaicking for viewing, 
one processed with conventional ISPs before the MobileNet network, and two other processed 
using \changed{jointly trained models} with and without Anscombe layer. As with the simulated data, 
conventional ISPs produce visually pleasing images by removing severe noise to a certain extent. 
However, fine details are lost in the process, leading to an incorrect 
classification result. The proposed joint stack does preserve and 
amplifies fine detail necessary for correct classification of the images. 

\subsection{Single-Image RAW Image Reconstruction in Low-Light for Human Viewing}\label{sec:imaging_results}
Next, we evaluate the proposed Anscombe network architecture when trained as an ISP 
replacement for human viewing. 
Specifically, we demonstrate joint demosaicking, 
denoising and tonemapping for human viewing on a single capture in low light, using the training and 
validation data set from~\cite{chen2018learning}. We employ the identical Anscombe 
network architecture from Sec.~\ref{sec:math} but, instead of concatenating this 
model with a higher-level domain-specific network, we minimize a loss $\higherloss$ 
formulated directly on the output image of the Anscombe network. 
This loss penalizes the difference between the prediction 
for a noisy observation and the corresponding 
clean long-exposure capture processed by a conventional ISP 
(with settings for normal lighting conditions). We use 
an $\ell_1$-loss after evaluating other 
alternative loss functions. 

\paragraph{\changed{Anscombe Networks also achieve state-of-the-art low-light performance for human viewing.}}
The results in Table~\ref{tab:low_light} show that our proposed 
model also obtains state-of-the-art performance for low-light image processing for human viewing. \changed{Our method outperforms the U-Net-based deep ISP~\cite{chen2018learning} qualitatively and quantitatively. 
We visualize RAW imaging results 
obtained by the evaluated methods in Fig.~\ref{fig:RAW_image_processing}}. 
\begin{table}[t!]
\vspace{-3pt}
  \centering
	\ra{1.0}
 	\resizebox{0.55\columnwidth}{!}{%
    {
	\begin{tabular}{rccccc}\toprule
		& PSNR & SSIM \\ \midrule
        Darktable ISP~\footnotemark  & 8.94 & 0.03 \\
		Chen et al.~\shortcite{chen2018learning} & 28.88 & 0.79  \\
		Proposed & \textbf{29.14} & \textbf{0.81} \\
		\bottomrule
	\end{tabular}
	} 	}
\vspace{3pt}
	\caption{Anscombe Networks for Single-Image Low-Light Photography 
	(w/o Classification). PSNR and SSIM comparison for Darktable ISP, 
	Chen et al.~\shortcite{chen2018learning}'s learned U-Net ISP, and 
	the proposed method, using the same training and test dataset 
	proposed by~\cite{chen2018learning}.\vspace{-20pt}}
	\label{tab:low_light}
\end{table}
\footnotetext{Traditional processing pipelines suffer also from severe color and white-balance artifacts in low-light such that quantiative results offer little insight. See Figure~\ref{fig:RAW_image_processing} for qualitative examples.}
In the presented low-light scenario, conventional ISPs fail due to the significant 
noise degradations affecting the RAW sensor readings. In particular, the darktable ISP produces severe 
chromatic artifacts in smooth image regions. Furthermore, fine details at object boundaries are severely 
distorted as a result of an edge-preserving denoising block. In contrast, the plain 
U-Net model proposed in~\cite{chen2018learning} produces visually pleasing images without 
chromatic artifact and free of residual noise. Chen et al.'s 
method also over-smooths image regions, i.e. noise is suppressed at the cost 
of texture loss. This behavior is particularly prevalent 
in areas with high intensity variations, around depth and illumination edges. 
The proposed Anscombe network model is tailored to intensity-dependent noise, 
and hence restores fine detail without over-smoothing or re-introducing 
residual noise. \changed{The results validate that Anscombe networks have the potential to be not only a domain-specific replacement for 
conventional general-purpose ISPs when considering non-traditional perception tasks 
but also when specialized to processing images for human viewing. 
We note that \emph{specialized domain-specific 
processing is the goal of the proposed approach}. We do not dismiss traditional 
ISPs when the downstream application is unknown but highlight 
the potential of domain-specific camera processing pipelines.}

\section{Mobile Prototype}
\label{sec:prototype}

We have implemented our joint low/high-level classification architecture on a mobile smartphone prototype along with a remote TensorFlow model server. The smartphone front-end application handles all dynamic camera control and the capture itself. 
While we rely on the hardware ISP for control of white-balance and auto-focus, we manually fix the exposure to ensure repeatable measurements with consistent signal-to-noise ratio. We use the Android Camera2 API for capture control and acquisition of the raw measurements. The captured raw date is transferred to a remote instance using TensorFlow's high-performance protocol buffer serving system, which then performs the inference on the transferred data. We use an Amazon Web Services P2.1x GPU instance to host the servables for our joint models, and baseline models for comparison. Fig.~\ref{fig:prototype} shows a photograph of the deployed application that classifies captures in the wild. 

We achieve an inference throughput of about $60$~FPS, while the vanilla MobileNet network performs at $80$~FPS under the same conditions. 
Note that \changed{this performance is achieved without any inference optimization or integer-quantization, which frameworks such as TensorRT offer}. We leave an efficient 
embedded hardware implementation as future work and  note that variants of 
the MobileNet architecture already achieve interactive 
framerates on mobile devices~\cite{mobilenet}.

\subsubsection{Ultra Low-light Classification}\label{sec:results_lowlight}

\begin{figure}[t]
	\centering
	\includegraphics[width=\linewidth]{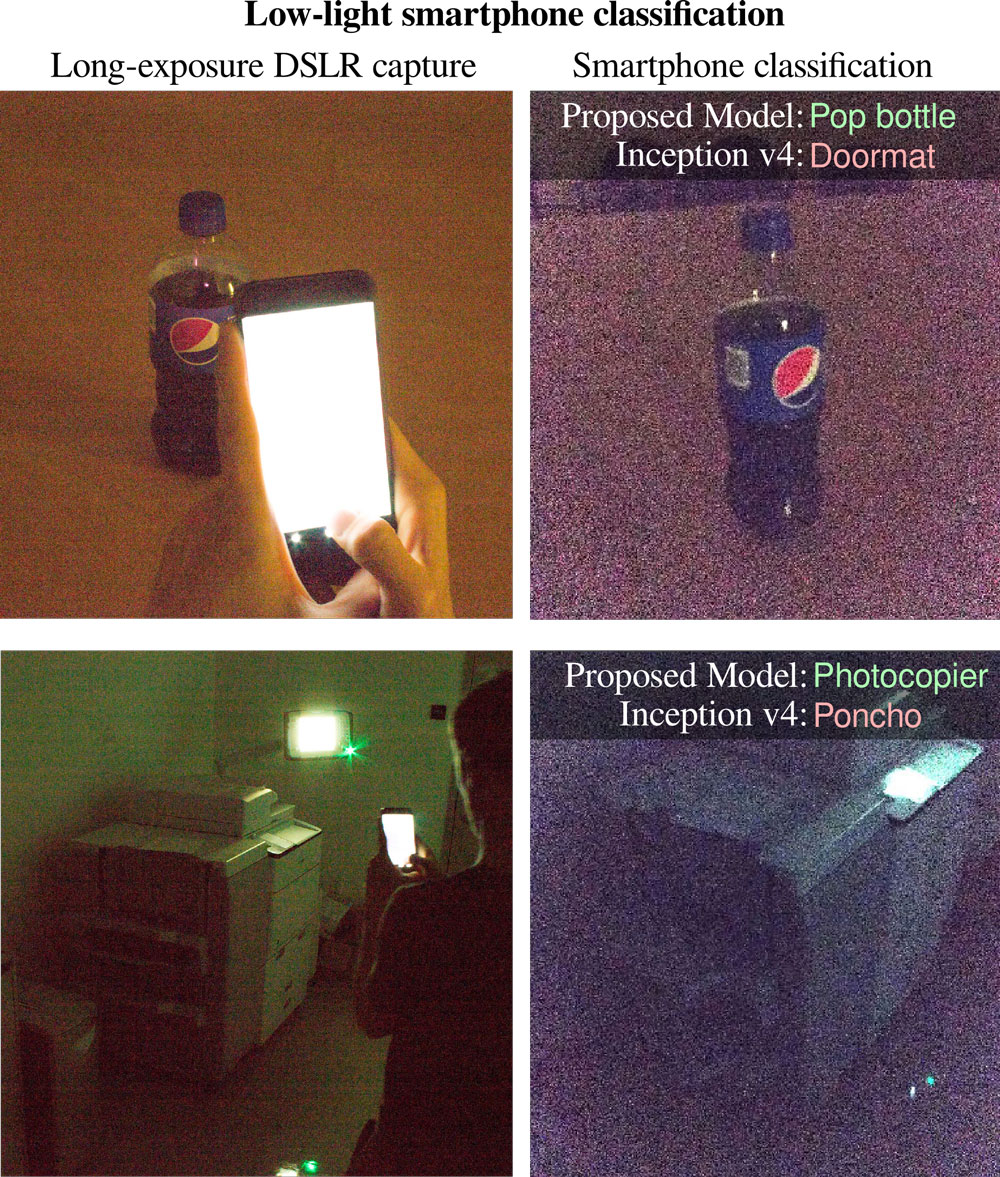}\vspace{-6pt}
	\caption{Extreme low-light cellphone classification. Two scenes acquired without any light sources other than the cellphone screen (and printer LCD screen). The left column shows scene captures acquired over a long 2 second exposure using a Canon Rebel T4 DSLR camera. Note that these are still severely degraded by noise. The right column shows the corresponding mobile capture, acquired over a $125$ms exposure, along with the classification label under these extreme conditions.}
	\label{fig:prototype}\vspace{-10pt}
\end{figure}
The mobile prototype performs classification tasks robustly even in extreme
low-light scenarios. Fig.~\ref{fig:prototype} shows two such challenging capture
scenarios along with classification results of the proposed and fine-tuned conventional MobileNet model.
Both scenes were captured in a closed room without windows or other sources of
ambient illumination. The only light sources present at the capture were the
phone screen's illumination and the photocopier's dim LCD screen light. The
scene captures shown in the left row of Fig.~\ref{fig:prototype} were captured
with a Canon Rebel T4 (f/4.5) with a long 2 second exposure at f/4.5. Note that
even these DSLR setup shots are severely degraded by noise due to the low scene
illumination. We acquired cellphone images with a long exposure of $125$ms
which, however, still allows for interactive frame-rates. The mobile prototype
correctly performs classification even in these extreme imaging scenarios, where
the from-scratch and fine-tuned MobileNet models fail. Please see the
supplemental video for additional low-light classification results in the wild.

\section{Discussion}
\label{sec:discussion}
In summary, we showed that the performance of conventional imaging and perception stacks, combining a high-quality ISP for human viewing with high-level networks trained on clean JPEG datasets, fails in low-light capture scenarios (and with optical off-axis aberrations of inexpensive mobile optics). Moreover, training classification architectures from scratch without any ISP outperforms sequential fine-tuned architectures that include an ISP, seemingly advocating for the removal of an ISP for higher-level image analysis tasks.

In this work, we investigated learned processing architectures that perform end-to-end image processing and perception jointly. The proposed Anscombe networks act as an ISP, using RAW color filter array data as input, and is flexible to transfer to different sensor architectures and capture settings without retraining or capture of new training datasets. However, by making the model end-to-end differentiable, the architecture can be trained jointly for a high-level loss function, achieving state-of-the-art performance both for RAW image processing for human viewing and perception tasks across light levels from ultra-low light to well-lit scenes.

We demonstrated that the proposed architecture makes imaging and perception robust to the extreme capture scenarios that can be commonly found in real-world imaging. We highlighted major qualitative differences between sequential approaches and our joint end-to-end approach by visualizing intermediate representations in the proposed architecture 
and the output of conventional pipelines algorithms. We demonstrated that Anscombe networks generalize across camera architectures, including different CFA patterns, optical systems and noise models, promising that analogue neural ISPs can be developed for other sensors modalities across computational imaging, such as time-of-flight cameras, multi-spectral cameras, and sensor fusion systems.

\paragraph{Limitations and Future Work}

While our proposed end-to-end model handles all the processing and image analysis after a RAW measurement has been acquired, a limitation of the method is that it does not address the dynamic control aspect of the capturing process, which is handled by the remaining trunk of the traditional ISP. 
\revised{Our proposed model then does not perform camera-control tasks, such as white-balance or auto-exposure. In the future, we plan to include auto-exposure and white-balance control in the proposed end-to-end model. These control tasks are particularly suited to include as image analysis feedback could severely affect the performance of these highly ill-posed problems.}

In the future, we will also expand the proposed architecture to model the camera optics and sensors as unknowns. Just as we optimized the full perception and imaging stack, we aim to optimize the optics, CFA pattern, and other elements of the imaging system for the given high-level vision task, effectively learning not only the processing but also the camera architecture itself.

%
%
\section{Conclusion}
\label{sec:conclusion}

In the future a large portion of the images taken by cameras and other imaging systems will be consumed by high-level perception stacks, not by humans.
We must reexamine the foundational assumptions of image processing in light of this momentous change.
Image reconstruction algorithms designed to produce visually pleasing images for
humans are not necessarily appropriate for a given perception task.
We have proposed one approach to redesigning low-level processing pipelines in an end-to-end optimization framework,
in a way that incorporates and benefits from knowledge of the physical image formation model and producing high-quality perceptually pleasing images when optimized for human-viewing.



\begin{acks}
\final{
Vincent Sitzmann was supported by a Stanford Graduate Fellowship in Science and Engineering. 
Gordon Wetzstein was supported by a National Science Foundation (NSF) CAREER award (IIS 1553333), 
a Sloan Fellowship, a PECASE from the ARO, and by the KAUST Office of Sponsored Research 
through the Visual Computing Center CCF grant.
Felix Heide was supported by an NSF CAREER Award (2047359).}
\end{acks}


\bibliographystyle{ACM-Reference-Format}
\bibliography{references}


\begin{thebibliography}{62}


\ifx \showCODEN    \undefined \def \showCODEN     #1{\unskip}     \fi
\ifx \showDOI      \undefined \def \showDOI       #1{#1}\fi
\ifx \showISBNx    \undefined \def \showISBNx     #1{\unskip}     \fi
\ifx \showISBNxiii \undefined \def \showISBNxiii  #1{\unskip}     \fi
\ifx \showISSN     \undefined \def \showISSN      #1{\unskip}     \fi
\ifx \showLCCN     \undefined \def \showLCCN      #1{\unskip}     \fi
\ifx \shownote     \undefined \def \shownote      #1{#1}          \fi
\ifx \showarticletitle \undefined \def \showarticletitle #1{#1}   \fi
\ifx \showURL      \undefined \def \showURL       {\relax}        \fi
\providecommand\bibfield[2]{#2}
\providecommand\bibinfo[2]{#2}
\providecommand\natexlab[1]{#1}
\providecommand\showeprint[2][]{arXiv:#2}

\bibitem[\protect\citeauthoryear{Agostinelli, Anderson, and Lee}{Agostinelli
  et~al\mbox{.}}{2013}]%
        {burges:13}
\bibfield{author}{\bibinfo{person}{F. Agostinelli}, \bibinfo{person}{M.
  Anderson}, {and} \bibinfo{person}{H. Lee}.} \bibinfo{year}{2013}\natexlab{}.
\newblock \showarticletitle{Adaptive Multi-Column Deep Neural Networks with
  Application to Robust Image Denoising}. In \bibinfo{booktitle}{\emph{Advances
  in Neural Information Processing Systems}},
  \bibfield{editor}{\bibinfo{person}{C.~Burges}, \bibinfo{person}{L.~Bottou},
  \bibinfo{person}{M.~Welling}, \bibinfo{person}{Z.~Ghahramani}, {and}
  \bibinfo{person}{K.~Weinberger}} (Eds.). \bibinfo{pages}{1493--1501}.
\newblock


\bibitem[\protect\citeauthoryear{Beck and Teboulle}{Beck and Teboulle}{2009a}]%
        {TB:09}
\bibfield{author}{\bibinfo{person}{A. Beck} {and} \bibinfo{person}{M.
  Teboulle}.} \bibinfo{year}{2009}\natexlab{a}.
\newblock \showarticletitle{Fast Gradient-Based Algorithms for Constrained
  Total Variation Image Denoising and Deblurring Problems}.
\newblock \bibinfo{journal}{\emph{IEEE Trans. Image Processing}}
  \bibinfo{volume}{18}, \bibinfo{number}{11} (\bibinfo{year}{2009}),
  \bibinfo{pages}{2419--2434}.
\newblock


\bibitem[\protect\citeauthoryear{Beck and Teboulle}{Beck and Teboulle}{2009b}]%
        {doi:10.1137/080716542}
\bibfield{author}{\bibinfo{person}{A. Beck} {and} \bibinfo{person}{M.
  Teboulle}.} \bibinfo{year}{2009}\natexlab{b}.
\newblock \showarticletitle{A Fast Iterative Shrinkage-Thresholding Algorithm
  for Linear Inverse Problems}.
\newblock \bibinfo{journal}{\emph{SIAM Journal on Imaging Sciences}}
  \bibinfo{volume}{2}, \bibinfo{number}{1} (\bibinfo{year}{2009}),
  \bibinfo{pages}{183--202}.
\newblock


\bibitem[\protect\citeauthoryear{Blasinski, Farrell, Lian, Liu, and
  Wandell}{Blasinski et~al\mbox{.}}{2018}]%
        {blasinski2018optimizing}
\bibfield{author}{\bibinfo{person}{Henryk Blasinski}, \bibinfo{person}{Joyce
  Farrell}, \bibinfo{person}{Trisha Lian}, \bibinfo{person}{Zhenyi Liu}, {and}
  \bibinfo{person}{Brian Wandell}.} \bibinfo{year}{2018}\natexlab{}.
\newblock \showarticletitle{Optimizing Image Acquisition Systems for Autonomous
  Driving}.
\newblock \bibinfo{journal}{\emph{Electronic Imaging}} \bibinfo{volume}{2018},
  \bibinfo{number}{5} (\bibinfo{year}{2018}), \bibinfo{pages}{1--7}.
\newblock


\bibitem[\protect\citeauthoryear{{Brooks}, {Mildenhall}, {Xue}, {Chen},
  {Sharlet}, and {Barron}}{{Brooks} et~al\mbox{.}}{2019}]%
        {Brooks:2019}
\bibfield{author}{\bibinfo{person}{T. {Brooks}}, \bibinfo{person}{B.
  {Mildenhall}}, \bibinfo{person}{T. {Xue}}, \bibinfo{person}{J. {Chen}},
  \bibinfo{person}{D. {Sharlet}}, {and} \bibinfo{person}{J.~T. {Barron}}.}
  \bibinfo{year}{2019}\natexlab{}.
\newblock \showarticletitle{Unprocessing Images for Learned Raw Denoising}. In
  \bibinfo{booktitle}{\emph{2019 IEEE/CVF Conference on Computer Vision and
  Pattern Recognition (CVPR)}}. \bibinfo{pages}{11028--11037}.
\newblock


\bibitem[\protect\citeauthoryear{Buades, Coll, and Morel}{Buades
  et~al\mbox{.}}{2005}]%
        {buades2005non}
\bibfield{author}{\bibinfo{person}{A. Buades}, \bibinfo{person}{B. Coll}, {and}
  \bibinfo{person}{J.-M. Morel}.} \bibinfo{year}{2005}\natexlab{}.
\newblock \showarticletitle{A non-local algorithm for image denoising}. In
  \bibinfo{booktitle}{\emph{Proc. IEEE CVPR}}, Vol.~\bibinfo{volume}{2}.
  \bibinfo{pages}{60--65}.
\newblock


\bibitem[\protect\citeauthoryear{Buckler, Jayasuriya, and Sampson}{Buckler
  et~al\mbox{.}}{2017}]%
        {Buckler2017}
\bibfield{author}{\bibinfo{person}{Mark Buckler}, \bibinfo{person}{Suren
  Jayasuriya}, {and} \bibinfo{person}{Adrian Sampson}.}
  \bibinfo{year}{2017}\natexlab{}.
\newblock \showarticletitle{Reconfiguring the Imaging Pipeline for Computer
  Vision}. In \bibinfo{booktitle}{\emph{IEEE International Conference on
  Computer Vision (ICCV)}}. \bibinfo{pages}{975--984}.
\newblock


\bibitem[\protect\citeauthoryear{Chambolle and Pock}{Chambolle and
  Pock}{2011}]%
        {chambolle2011first}
\bibfield{author}{\bibinfo{person}{Antonin Chambolle} {and}
  \bibinfo{person}{Thomas Pock}.} \bibinfo{year}{2011}\natexlab{}.
\newblock \showarticletitle{A first-order primal-dual algorithm for convex
  problems with applications to imaging}.
\newblock \bibinfo{journal}{\emph{Journal of Mathematical Imaging and Vision}}
  \bibinfo{volume}{40}, \bibinfo{number}{1} (\bibinfo{year}{2011}),
  \bibinfo{pages}{120--145}.
\newblock


\bibitem[\protect\citeauthoryear{{Chen}, {Chen}, {Xu}, and {Koltun}}{{Chen}
  et~al\mbox{.}}{2018}]%
        {chen2018learning}
\bibfield{author}{\bibinfo{person}{C. {Chen}}, \bibinfo{person}{Q. {Chen}},
  \bibinfo{person}{J. {Xu}}, {and} \bibinfo{person}{V. {Koltun}}.}
  \bibinfo{year}{2018}\natexlab{}.
\newblock \showarticletitle{{Learning to See in the Dark}}.
\newblock \bibinfo{journal}{\emph{ArXiv e-prints}} (\bibinfo{date}{May}
  \bibinfo{year}{2018}).
\newblock
\showeprint[arxiv]{1805.01934}


\bibitem[\protect\citeauthoryear{Chen, Li, and Srihari}{Chen
  et~al\mbox{.}}{2016}]%
        {chen2016joint}
\bibfield{author}{\bibinfo{person}{G. Chen}, \bibinfo{person}{Y. Li}, {and}
  \bibinfo{person}{S. Srihari}.} \bibinfo{year}{2016}\natexlab{}.
\newblock \showarticletitle{Joint visual denoising and classification using
  deep learning}. In \bibinfo{booktitle}{\emph{Proceedings of the IEEE
  International Conference on Image Processing}}. \bibinfo{pages}{3673--3677}.
\newblock


\bibitem[\protect\citeauthoryear{da~Costa, Contato, Nazare, Neto, and
  Ponti}{da~Costa et~al\mbox{.}}{2016}]%
        {da2016empirical}
\bibfield{author}{\bibinfo{person}{G. da Costa}, \bibinfo{person}{W. Contato},
  \bibinfo{person}{T. Nazare}, \bibinfo{person}{J. Neto}, {and}
  \bibinfo{person}{M. Ponti}.} \bibinfo{year}{2016}\natexlab{}.
\newblock \showarticletitle{An empirical study on the effects of different
  types of noise in image classification tasks}.
\newblock \bibinfo{journal}{\emph{arXiv preprint arXiv:1609.02781}}
  (\bibinfo{year}{2016}).
\newblock


\bibitem[\protect\citeauthoryear{Dabov, Foi, Katkovnik, and Egiazarian}{Dabov
  et~al\mbox{.}}{2007}]%
        {dabov2007image}
\bibfield{author}{\bibinfo{person}{K. Dabov}, \bibinfo{person}{A. Foi},
  \bibinfo{person}{V. Katkovnik}, {and} \bibinfo{person}{K. Egiazarian}.}
  \bibinfo{year}{2007}\natexlab{}.
\newblock \showarticletitle{Image denoising by sparse 3-D transform-domain
  collaborative filtering}.
\newblock \bibinfo{journal}{\emph{IEEE Trans. Image Processing}}
  \bibinfo{volume}{16}, \bibinfo{number}{8} (\bibinfo{year}{2007}),
  \bibinfo{pages}{2080--2095}.
\newblock


\bibitem[\protect\citeauthoryear{darktable}{darktable}{2018}]%
        {darktable}
\bibfield{author}{\bibinfo{person}{darktable}.}
  \bibinfo{year}{2018}\natexlab{}.
\newblock \bibinfo{title}{darktable version 2.4.3}.
\newblock \bibinfo{howpublished}{\url{https://www.darktable.org}}.
\newblock


\bibitem[\protect\citeauthoryear{Deng, Dong, Socher, Li, Li, and Fei-Fei}{Deng
  et~al\mbox{.}}{2009}]%
        {deng2009imagenet}
\bibfield{author}{\bibinfo{person}{J. Deng}, \bibinfo{person}{W. Dong},
  \bibinfo{person}{R. Socher}, \bibinfo{person}{L.-J. Li}, \bibinfo{person}{K.
  Li}, {and} \bibinfo{person}{L. Fei-Fei}.} \bibinfo{year}{2009}\natexlab{}.
\newblock \showarticletitle{Imagenet: A large-scale hierarchical image
  database}. In \bibinfo{booktitle}{\emph{Computer Vision and Pattern
  Recognition, 2009. CVPR 2009. IEEE Conference on}}. IEEE,
  \bibinfo{pages}{248--255}.
\newblock


\bibitem[\protect\citeauthoryear{Diamond, Sitzmann, Heide, and
  Wetzstein}{Diamond et~al\mbox{.}}{2017}]%
        {odp:17}
\bibfield{author}{\bibinfo{person}{S. Diamond}, \bibinfo{person}{V. Sitzmann},
  \bibinfo{person}{F. Heide}, {and} \bibinfo{person}{G. Wetzstein}.}
  \bibinfo{year}{2017}\natexlab{}.
\newblock \showarticletitle{Unrolled Optimization with Deep Priors}.
\newblock \bibinfo{journal}{\emph{arXiv preprint}} (\bibinfo{year}{2017}).
\newblock


\bibitem[\protect\citeauthoryear{Dodge and Karam}{Dodge and Karam}{2016}]%
        {dodge2016understanding}
\bibfield{author}{\bibinfo{person}{S. Dodge} {and} \bibinfo{person}{L. Karam}.}
  \bibinfo{year}{2016}\natexlab{}.
\newblock \showarticletitle{Understanding how image quality affects deep neural
  networks}. In \bibinfo{booktitle}{\emph{International Conference on Quality
  of Multimedia Experience}}. \bibinfo{pages}{1--6}.
\newblock


\bibitem[\protect\citeauthoryear{Foi}{Foi}{2009}]%
        {foi2009clipped}
\bibfield{author}{\bibinfo{person}{A. Foi}.} \bibinfo{year}{2009}\natexlab{}.
\newblock \showarticletitle{Clipped noisy images: Heteroskedastic modeling and
  practical denoising}.
\newblock \bibinfo{journal}{\emph{Signal Processing}} \bibinfo{volume}{89},
  \bibinfo{number}{12} (\bibinfo{year}{2009}), \bibinfo{pages}{2609--2629}.
\newblock


\bibitem[\protect\citeauthoryear{Foi and Makitalo}{Foi and Makitalo}{2013}]%
        {FM:13}
\bibfield{author}{\bibinfo{person}{A. Foi} {and} \bibinfo{person}{M.
  Makitalo}.} \bibinfo{year}{2013}\natexlab{}.
\newblock \showarticletitle{Optimal inversion of the generalized {A}nscombe
  transformation for {P}oisson-{G}aussian noise}.
\newblock \bibinfo{journal}{\emph{IEEE Trans. Image Process.}}
  \bibinfo{volume}{22}, \bibinfo{number}{1} (\bibinfo{year}{2013}),
  \bibinfo{pages}{91--103}.
\newblock


\bibitem[\protect\citeauthoryear{Foi, Trimeche, Katkovnik, and Egiazarian}{Foi
  et~al\mbox{.}}{2008}]%
        {foi08}
\bibfield{author}{\bibinfo{person}{A. Foi}, \bibinfo{person}{M. Trimeche},
  \bibinfo{person}{V. Katkovnik}, {and} \bibinfo{person}{K. Egiazarian}.}
  \bibinfo{year}{2008}\natexlab{}.
\newblock \showarticletitle{Practical {P}oissonian-{G}aussian noise modeling
  and fitting for single-image raw-data}.
\newblock \bibinfo{journal}{\emph{IEEE Trans. Image Process.}}
  \bibinfo{volume}{17}, \bibinfo{number}{10} (\bibinfo{year}{2008}),
  \bibinfo{pages}{1737--1754}.
\newblock


\bibitem[\protect\citeauthoryear{Ganin and Lempitsky}{Ganin and
  Lempitsky}{2015}]%
        {ganin2015unsupervised}
\bibfield{author}{\bibinfo{person}{Y. Ganin} {and} \bibinfo{person}{V.
  Lempitsky}.} \bibinfo{year}{2015}\natexlab{}.
\newblock \showarticletitle{Unsupervised domain adaptation by backpropagation}.
  In \bibinfo{booktitle}{\emph{International Conference on Machine Learning}}.
  \bibinfo{pages}{1180--1189}.
\newblock


\bibitem[\protect\citeauthoryear{Gharbi, Chaurasia, Paris, and Durand}{Gharbi
  et~al\mbox{.}}{2016}]%
        {gharbi2016deep}
\bibfield{author}{\bibinfo{person}{M. Gharbi}, \bibinfo{person}{G. Chaurasia},
  \bibinfo{person}{S. Paris}, {and} \bibinfo{person}{F. Durand}.}
  \bibinfo{year}{2016}\natexlab{}.
\newblock \showarticletitle{Deep joint demosaicking and denoising}.
\newblock \bibinfo{journal}{\emph{ACM Transactions on Graphics (TOG)}}
  \bibinfo{volume}{35}, \bibinfo{number}{6} (\bibinfo{year}{2016}),
  \bibinfo{pages}{191}.
\newblock


\bibitem[\protect\citeauthoryear{Gharbi, Chen, Barron, Hasinoff, and
  Durand}{Gharbi et~al\mbox{.}}{2017}]%
        {GharbiSIGGRAPH2017}
\bibfield{author}{\bibinfo{person}{M. Gharbi}, \bibinfo{person}{J. Chen},
  \bibinfo{person}{J. Barron}, \bibinfo{person}{S. Hasinoff}, {and}
  \bibinfo{person}{F. Durand}.} \bibinfo{year}{2017}\natexlab{}.
\newblock \showarticletitle{Deep Bilateral Learning for Real-Time Image
  Enhancement}.
\newblock \bibinfo{journal}{\emph{SIGGRAPH}} (\bibinfo{year}{2017}).
\newblock


\bibitem[\protect\citeauthoryear{Glowinski and Marroco}{Glowinski and
  Marroco}{1975}]%
        {glowinski1975approximation}
\bibfield{author}{\bibinfo{person}{R. Glowinski} {and} \bibinfo{person}{A.
  Marroco}.} \bibinfo{year}{1975}\natexlab{}.
\newblock \showarticletitle{Sur l'approximation, par {\'e}l{\'e}ments finis
  d'ordre un, et la r{\'e}solution, par p{\'e}nalisation-dualit{\'e} d'une
  classe de probl{\`e}mes de Dirichlet non lin{\'e}aires}.
\newblock \bibinfo{journal}{\emph{Revue fran{\c{c}}aise d'automatique,
  informatique, recherche op{\'e}rationnelle. Analyse num{\'e}rique}}
  \bibinfo{volume}{9}, \bibinfo{number}{2} (\bibinfo{year}{1975}),
  \bibinfo{pages}{41--76}.
\newblock


\bibitem[\protect\citeauthoryear{Hasinoff, Sharlet, Geiss, Adams, Barron,
  Kainz, Chen, and Levoy}{Hasinoff et~al\mbox{.}}{2016}]%
        {hasinoff2016}
\bibfield{author}{\bibinfo{person}{S. Hasinoff}, \bibinfo{person}{D. Sharlet},
  \bibinfo{person}{R. Geiss}, \bibinfo{person}{A. Adams}, \bibinfo{person}{J.
  Barron}, \bibinfo{person}{F. Kainz}, \bibinfo{person}{J. Chen}, {and}
  \bibinfo{person}{M. Levoy}.} \bibinfo{year}{2016}\natexlab{}.
\newblock \showarticletitle{Burst Photography for High Dynamic Range and
  Low-light Imaging on Mobile Cameras}.
\newblock \bibinfo{journal}{\emph{ACM Trans. Graph.}} \bibinfo{volume}{35},
  \bibinfo{number}{6}, Article \bibinfo{articleno}{192} (\bibinfo{year}{2016}),
  \bibinfo{numpages}{12}~pages.
\newblock


\bibitem[\protect\citeauthoryear{Hegarty, Brunhaver, DeVito, Ragan-Kelley,
  Cohen, Bell, Vasilyev, Horowitz, and Hanrahan}{Hegarty et~al\mbox{.}}{2014}]%
        {Hegarty:2014}
\bibfield{author}{\bibinfo{person}{James Hegarty}, \bibinfo{person}{John
  Brunhaver}, \bibinfo{person}{Zachary DeVito}, \bibinfo{person}{Jonathan
  Ragan-Kelley}, \bibinfo{person}{Noy Cohen}, \bibinfo{person}{Steven Bell},
  \bibinfo{person}{Artem Vasilyev}, \bibinfo{person}{Mark Horowitz}, {and}
  \bibinfo{person}{Pat Hanrahan}.} \bibinfo{year}{2014}\natexlab{}.
\newblock \showarticletitle{Darkroom: Compiling High-level Image Processing
  Code into Hardware Pipelines}.
\newblock \bibinfo{journal}{\emph{ACM Trans. Graph. (SIGGRAPH)}}
  \bibinfo{volume}{33}, \bibinfo{number}{4} (\bibinfo{year}{2014}).
\newblock


\bibitem[\protect\citeauthoryear{Heide, Steinberger, Tsai, Rouf, Pajak, Reddy,
  Gallo, Liu, Heidrich, Egiazarian, Kautz, and Pulli}{Heide
  et~al\mbox{.}}{2014}]%
        {heide2014flexisp}
\bibfield{author}{\bibinfo{person}{F. Heide}, \bibinfo{person}{M. Steinberger},
  \bibinfo{person}{Y.-T. Tsai}, \bibinfo{person}{M. Rouf}, \bibinfo{person}{D.
  Pajak}, \bibinfo{person}{D. Reddy}, \bibinfo{person}{O. Gallo},
  \bibinfo{person}{J. Liu}, \bibinfo{person}{W. Heidrich}, \bibinfo{person}{K.
  Egiazarian}, \bibinfo{person}{J. Kautz}, {and} \bibinfo{person}{K. Pulli}.}
  \bibinfo{year}{2014}\natexlab{}.
\newblock \showarticletitle{Flex{ISP}: A flexible camera image processing
  framework}.
\newblock \bibinfo{journal}{\emph{ACM Trans. Graph. (SIGGRAPH Asia)}}
  \bibinfo{volume}{33}, \bibinfo{number}{6} (\bibinfo{year}{2014}).
\newblock


\bibitem[\protect\citeauthoryear{Hoffman, Tzeng, Park, Zhu, Isola, Saenko,
  Efros, and Darrell}{Hoffman et~al\mbox{.}}{2017}]%
        {hoffman2017cycada}
\bibfield{author}{\bibinfo{person}{Judy Hoffman}, \bibinfo{person}{Eric Tzeng},
  \bibinfo{person}{Taesung Park}, \bibinfo{person}{Jun-Yan Zhu},
  \bibinfo{person}{Phillip Isola}, \bibinfo{person}{Kate Saenko},
  \bibinfo{person}{Alexei~A Efros}, {and} \bibinfo{person}{Trevor Darrell}.}
  \bibinfo{year}{2017}\natexlab{}.
\newblock \showarticletitle{Cycada: Cycle-consistent adversarial domain
  adaptation}.
\newblock \bibinfo{journal}{\emph{arXiv preprint arXiv:1711.03213}}
  (\bibinfo{year}{2017}).
\newblock


\bibitem[\protect\citeauthoryear{Howard, Zhu, Chen, Kalenichenko, Wang, Weyand,
  Andreetto, and Adam}{Howard et~al\mbox{.}}{2017}]%
        {mobilenet}
\bibfield{author}{\bibinfo{person}{Andrew~G. Howard}, \bibinfo{person}{Menglong
  Zhu}, \bibinfo{person}{Bo Chen}, \bibinfo{person}{Dmitry Kalenichenko},
  \bibinfo{person}{Weijun Wang}, \bibinfo{person}{Tobias Weyand},
  \bibinfo{person}{Marco Andreetto}, {and} \bibinfo{person}{Hartwig Adam}.}
  \bibinfo{year}{2017}\natexlab{}.
\newblock \showarticletitle{MobileNets: Efficient Convolutional Neural Networks
  for Mobile Vision Applications}.
\newblock \bibinfo{journal}{\emph{CoRR}}  \bibinfo{volume}{abs/1704.04861}
  (\bibinfo{year}{2017}).
\newblock
\showeprint[arxiv]{1704.04861}
\urldef\tempurl%
\url{http://arxiv.org/abs/1704.04861}
\showURL{%
\tempurl}


\bibitem[\protect\citeauthoryear{Ioffe and Szegedy}{Ioffe and Szegedy}{2015}]%
        {ioffe2015batch}
\bibfield{author}{\bibinfo{person}{Sergey Ioffe} {and}
  \bibinfo{person}{Christian Szegedy}.} \bibinfo{year}{2015}\natexlab{}.
\newblock \showarticletitle{Batch normalization: Accelerating deep network
  training by reducing internal covariate shift}. In
  \bibinfo{booktitle}{\emph{International Conference on Machine Learning}}.
  \bibinfo{pages}{448--456}.
\newblock


\bibitem[\protect\citeauthoryear{{ISO 12233:2014}}{{ISO 12233:2014}}{2014}]%
        {iso12233}
{ISO 12233:2014} \bibinfo{year}{{2014}}\natexlab{}.
\newblock \bibinfo{title}{{ISO 12233:2014 Photography -- Electronic still
  picture imaging -- Resolution and spatial frequency responses}}.
\newblock
\newblock


\bibitem[\protect\citeauthoryear{Jalalvand, Neve, de~Walle, and
  Martens}{Jalalvand et~al\mbox{.}}{2016}]%
        {JNWM:16}
\bibfield{author}{\bibinfo{person}{A. Jalalvand}, \bibinfo{person}{W.~De Neve},
  \bibinfo{person}{R.~Van de Walle}, {and} \bibinfo{person}{J. Martens}.}
  \bibinfo{year}{2016}\natexlab{}.
\newblock \showarticletitle{Towards using Reservoir Computing Networks for
  noise-robust image recognition}. In \bibinfo{booktitle}{\emph{Proceedings of
  the International Joint Conference on Neural Networks}}.
  \bibinfo{pages}{1666--1672}.
\newblock


\bibitem[\protect\citeauthoryear{Jaroensri, Biscarrat, Aittala, and
  Durand}{Jaroensri et~al\mbox{.}}{2019}]%
        {Jaroensri:2019}
\bibfield{author}{\bibinfo{person}{Ronnachai Jaroensri},
  \bibinfo{person}{Camille Biscarrat}, \bibinfo{person}{Miika Aittala}, {and}
  \bibinfo{person}{Fr{\'e}do Durand}.} \bibinfo{year}{2019}\natexlab{}.
\newblock \showarticletitle{Generating Training Data for Denoising Real RGB
  Images via Camera Pipeline Simulation}.
\newblock \bibinfo{journal}{\emph{ArXiv}}  \bibinfo{volume}{abs/1904.08825}
  (\bibinfo{year}{2019}).
\newblock


\bibitem[\protect\citeauthoryear{Jin, Phillips, Farnand, Belska, Tran, Chang,
  Wang, and Tseng}{Jin et~al\mbox{.}}{2017}]%
        {jin2017towards}
\bibfield{author}{\bibinfo{person}{E. Jin}, \bibinfo{person}{J. Phillips},
  \bibinfo{person}{S. Farnand}, \bibinfo{person}{M. Belska},
  \bibinfo{person}{V. Tran}, \bibinfo{person}{E. Chang}, \bibinfo{person}{Y.
  Wang}, {and} \bibinfo{person}{B. Tseng}.} \bibinfo{year}{2017}\natexlab{}.
\newblock \showarticletitle{Towards the Development of the IEEE P1858 CPIQ
  Standard--A validation study}.
\newblock \bibinfo{journal}{\emph{Electronic Imaging}} \bibinfo{volume}{2017},
  \bibinfo{number}{12} (\bibinfo{year}{2017}), \bibinfo{pages}{88--94}.
\newblock


\bibitem[\protect\citeauthoryear{Johnson, Alahi, and Fei-Fei}{Johnson
  et~al\mbox{.}}{2016}]%
        {johnson2016perceptual}
\bibfield{author}{\bibinfo{person}{Justin Johnson}, \bibinfo{person}{Alexandre
  Alahi}, {and} \bibinfo{person}{Li Fei-Fei}.} \bibinfo{year}{2016}\natexlab{}.
\newblock \showarticletitle{Perceptual losses for real-time style transfer and
  super-resolution}. In \bibinfo{booktitle}{\emph{European Conference on
  Computer Vision}}. Springer, \bibinfo{pages}{694--711}.
\newblock


\bibitem[\protect\citeauthoryear{Karahan, Yildirum, Kirtac, Rende, Butun, and
  Ekenel}{Karahan et~al\mbox{.}}{2016}]%
        {karahan:16}
\bibfield{author}{\bibinfo{person}{S. Karahan}, \bibinfo{person}{M. Yildirum},
  \bibinfo{person}{K. Kirtac}, \bibinfo{person}{F. Rende}, \bibinfo{person}{G.
  Butun}, {and} \bibinfo{person}{H. Ekenel}.} \bibinfo{year}{2016}\natexlab{}.
\newblock \showarticletitle{How Image Degradations Affect Deep CNN-Based Face
  Recognition?}. In \bibinfo{booktitle}{\emph{International Conference of the
  Biometrics Special Interest Group}}. \bibinfo{pages}{1--5}.
\newblock


\bibitem[\protect\citeauthoryear{Liang, Cai, Cao, and Zhang}{Liang
  et~al\mbox{.}}{2019}]%
        {Liang:2019}
\bibfield{author}{\bibinfo{person}{Zhetong Liang}, \bibinfo{person}{Jianrui
  Cai}, \bibinfo{person}{Zisheng Cao}, {and} \bibinfo{person}{Lei Zhang}.}
  \bibinfo{year}{2019}\natexlab{}.
\newblock \showarticletitle{CameraNet: {A} Two-Stage Framework for Effective
  Camera {ISP} Learning}.
\newblock \bibinfo{journal}{\emph{CoRR}}  \bibinfo{volume}{abs/1908.01481}
  (\bibinfo{year}{2019}).
\newblock
\showeprint[arxiv]{1908.01481}
\urldef\tempurl%
\url{http://arxiv.org/abs/1908.01481}
\showURL{%
\tempurl}


\bibitem[\protect\citeauthoryear{Liba, Murthy, Tsai, Brooks, Xue, Karnad, He,
  Barron, Sharlet, Geiss, Hasinoff, Pritch, and Levoy}{Liba
  et~al\mbox{.}}{2019}]%
        {Liba:2019}
\bibfield{author}{\bibinfo{person}{Orly Liba}, \bibinfo{person}{Kiran Murthy},
  \bibinfo{person}{Yun-Ta Tsai}, \bibinfo{person}{Tim Brooks},
  \bibinfo{person}{Tianfan Xue}, \bibinfo{person}{Nikhil Karnad},
  \bibinfo{person}{Qiurui He}, \bibinfo{person}{Jonathan~T. Barron},
  \bibinfo{person}{Dillon Sharlet}, \bibinfo{person}{Ryan Geiss},
  \bibinfo{person}{Samuel~W. Hasinoff}, \bibinfo{person}{Yael Pritch}, {and}
  \bibinfo{person}{Marc Levoy}.} \bibinfo{year}{2019}\natexlab{}.
\newblock \showarticletitle{Handheld Mobile Photography in Very Low Light}.
\newblock \bibinfo{journal}{\emph{ACM Trans. Graph.}} \bibinfo{volume}{38},
  \bibinfo{number}{6}, Article \bibinfo{articleno}{164} (\bibinfo{date}{Nov.}
  \bibinfo{year}{2019}), \bibinfo{numpages}{16}~pages.
\newblock
\showISSN{0730-0301}
\urldef\tempurl%
\url{https://doi.org/10.1145/3355089.3356508}
\showDOI{\tempurl}


\bibitem[\protect\citeauthoryear{Lin, Maire, Belongie, Hays, Perona, Ramanan,
  Doll{\'a}r, and Zitnick}{Lin et~al\mbox{.}}{2014}]%
        {lin2014microsoft}
\bibfield{author}{\bibinfo{person}{Tsung-Yi Lin}, \bibinfo{person}{Michael
  Maire}, \bibinfo{person}{Serge Belongie}, \bibinfo{person}{James Hays},
  \bibinfo{person}{Pietro Perona}, \bibinfo{person}{Deva Ramanan},
  \bibinfo{person}{Piotr Doll{\'a}r}, {and} \bibinfo{person}{C~Lawrence
  Zitnick}.} \bibinfo{year}{2014}\natexlab{}.
\newblock \showarticletitle{Microsoft coco: Common objects in context}. In
  \bibinfo{booktitle}{\emph{European conference on computer vision}}. Springer,
  \bibinfo{pages}{740--755}.
\newblock


\bibitem[\protect\citeauthoryear{Long, Cao, Wang, and Jordan}{Long
  et~al\mbox{.}}{2015}]%
        {long2015learning}
\bibfield{author}{\bibinfo{person}{M. Long}, \bibinfo{person}{Y. Cao},
  \bibinfo{person}{J. Wang}, {and} \bibinfo{person}{M. Jordan}.}
  \bibinfo{year}{2015}\natexlab{}.
\newblock \showarticletitle{Learning transferable features with deep adaptation
  networks}. In \bibinfo{booktitle}{\emph{International Conference on Machine
  Learning}}. \bibinfo{pages}{97--105}.
\newblock


\bibitem[\protect\citeauthoryear{Moloney, Barry, Richmond, Connor, Brick, and
  Donohoe}{Moloney et~al\mbox{.}}{2014}]%
        {moloney2014myriad}
\bibfield{author}{\bibinfo{person}{D. Moloney}, \bibinfo{person}{B. Barry},
  \bibinfo{person}{R. Richmond}, \bibinfo{person}{F. Connor},
  \bibinfo{person}{C. Brick}, {and} \bibinfo{person}{D. Donohoe}.}
  \bibinfo{year}{2014}\natexlab{}.
\newblock \showarticletitle{Myriad 2: Eye of the computational vision storm}.
  In \bibinfo{booktitle}{\emph{Hot Chips 26 Symposium (HCS), 2014 IEEE}}. IEEE,
  \bibinfo{pages}{1--18}.
\newblock


\bibitem[\protect\citeauthoryear{Mosleh, Green, Onzon, Begin, and
  Pierre~Langlois}{Mosleh et~al\mbox{.}}{2015}]%
        {Mosleh_2015_CVPR}
\bibfield{author}{\bibinfo{person}{A. Mosleh}, \bibinfo{person}{P. Green},
  \bibinfo{person}{E. Onzon}, \bibinfo{person}{I. Begin}, {and}
  \bibinfo{person}{J.M. Pierre~Langlois}.} \bibinfo{year}{2015}\natexlab{}.
\newblock \showarticletitle{Camera Intrinsic Blur Kernel Estimation: A Reliable
  Framework}. In \bibinfo{booktitle}{\emph{The IEEE Conference on Computer
  Vision and Pattern Recognition (CVPR)}}.
\newblock


\bibitem[\protect\citeauthoryear{MT9P111}{MT9P111}{2015}]%
        {aptinaMT9P111}
\bibfield{author}{\bibinfo{person}{ON~Semi MT9P111}.}
  \bibinfo{year}{2015}\natexlab{}.
\newblock \bibinfo{title}{MT9P111: 1/4-Inch 5 Mp System-On-A-Chip (SOC) CMOS
  Digital Image Sensor}.
\newblock
  \bibinfo{howpublished}{\url{http://www.onsemi.com/pub/Collateral/MT9P111-D.PDF}}.
\newblock


\bibitem[\protect\citeauthoryear{Parikh and Boyd}{Parikh and Boyd}{2013}]%
        {parikh2013proximal}
\bibfield{author}{\bibinfo{person}{N. Parikh} {and} \bibinfo{person}{S. Boyd}.}
  \bibinfo{year}{2013}\natexlab{}.
\newblock \showarticletitle{Proximal algorithms}.
\newblock \bibinfo{journal}{\emph{Foundations and Trends in Optimization}}
  \bibinfo{volume}{1}, \bibinfo{number}{3} (\bibinfo{year}{2013}),
  \bibinfo{pages}{123--231}.
\newblock


\bibitem[\protect\citeauthoryear{Phillips and Eliasson}{Phillips and
  Eliasson}{2018}]%
        {phillips2018camera}
\bibfield{author}{\bibinfo{person}{Jonathan~B Phillips} {and}
  \bibinfo{person}{Henrik Eliasson}.} \bibinfo{year}{2018}\natexlab{}.
\newblock \bibinfo{booktitle}{\emph{Camera Image Quality Benchmarking}}.
\newblock \bibinfo{publisher}{John Wiley \& Sons}.
\newblock


\bibitem[\protect\citeauthoryear{Plotz and Roth}{Plotz and Roth}{2017}]%
        {Plotz_2017_CVPR}
\bibfield{author}{\bibinfo{person}{Tobias Plotz} {and} \bibinfo{person}{Stefan
  Roth}.} \bibinfo{year}{2017}\natexlab{}.
\newblock \showarticletitle{Benchmarking Denoising Algorithms With Real
  Photographs}. In \bibinfo{booktitle}{\emph{The IEEE Conference on Computer
  Vision and Pattern Recognition (CVPR)}}.
\newblock


\bibitem[\protect\citeauthoryear{Ramanath, Snyder, Yoo, and Drew}{Ramanath
  et~al\mbox{.}}{2005a}]%
        {ram:05}
\bibfield{author}{\bibinfo{person}{R. Ramanath}, \bibinfo{person}{W. Snyder},
  \bibinfo{person}{Y. Yoo}, {and} \bibinfo{person}{M. Drew}.}
  \bibinfo{year}{2005}\natexlab{a}.
\newblock \showarticletitle{Color image processing pipeline in digital still
  cameras}.
\newblock \bibinfo{journal}{\emph{IEEE Signal Processing Magazine}}
  \bibinfo{volume}{22}, \bibinfo{number}{1} (\bibinfo{year}{2005}),
  \bibinfo{pages}{34--43}.
\newblock


\bibitem[\protect\citeauthoryear{Ramanath, Snyder, Yoo, and Drew}{Ramanath
  et~al\mbox{.}}{2005b}]%
        {ramanath2005color}
\bibfield{author}{\bibinfo{person}{Rajeev Ramanath}, \bibinfo{person}{Wesley~E
  Snyder}, \bibinfo{person}{Youngjun Yoo}, {and} \bibinfo{person}{Mark~S
  Drew}.} \bibinfo{year}{2005}\natexlab{b}.
\newblock \showarticletitle{Color image processing pipeline}.
\newblock \bibinfo{journal}{\emph{IEEE Signal Processing Magazine}}
  \bibinfo{volume}{22}, \bibinfo{number}{1} (\bibinfo{year}{2005}),
  \bibinfo{pages}{34--43}.
\newblock


\bibitem[\protect\citeauthoryear{Ronneberger, Fischer, and Brox}{Ronneberger
  et~al\mbox{.}}{2015}]%
        {unetRonneberger}
\bibfield{author}{\bibinfo{person}{Olaf Ronneberger}, \bibinfo{person}{Philipp
  Fischer}, {and} \bibinfo{person}{Thomas Brox}.}
  \bibinfo{year}{2015}\natexlab{}.
\newblock \showarticletitle{U-Net: Convolutional Networks for Biomedical Image
  Segmentation}.
\newblock \bibinfo{journal}{\emph{CoRR}}  \bibinfo{volume}{abs/1505.04597}
  (\bibinfo{year}{2015}).
\newblock
\showeprint[arxiv]{1505.04597}
\urldef\tempurl%
\url{http://arxiv.org/abs/1505.04597}
\showURL{%
\tempurl}


\bibitem[\protect\citeauthoryear{{Sandler}, {Howard}, {Zhu}, {Zhmoginov}, and
  {Chen}}{{Sandler} et~al\mbox{.}}{2018}]%
        {Sandler:2018}
\bibfield{author}{\bibinfo{person}{M. {Sandler}}, \bibinfo{person}{A.
  {Howard}}, \bibinfo{person}{M. {Zhu}}, \bibinfo{person}{A. {Zhmoginov}},
  {and} \bibinfo{person}{L. {Chen}}.} \bibinfo{year}{2018}\natexlab{}.
\newblock \showarticletitle{MobileNetV2: Inverted Residuals and Linear
  Bottlenecks}. In \bibinfo{booktitle}{\emph{2018 IEEE/CVF Conference on
  Computer Vision and Pattern Recognition}}. \bibinfo{pages}{4510--4520}.
\newblock


\bibitem[\protect\citeauthoryear{{Schwartz}, {Giryes}, and
  {Bronstein}}{{Schwartz} et~al\mbox{.}}{2019}]%
        {Schwartz:2019}
\bibfield{author}{\bibinfo{person}{E. {Schwartz}}, \bibinfo{person}{R.
  {Giryes}}, {and} \bibinfo{person}{A.~M. {Bronstein}}.}
  \bibinfo{year}{2019}\natexlab{}.
\newblock \showarticletitle{DeepISP: Toward Learning an End-to-End Image
  Processing Pipeline}.
\newblock \bibinfo{journal}{\emph{IEEE Transactions on Image Processing}}
  \bibinfo{volume}{28}, \bibinfo{number}{2} (\bibinfo{year}{2019}),
  \bibinfo{pages}{912--923}.
\newblock


\bibitem[\protect\citeauthoryear{Shao, Yan, Li, and Liu}{Shao
  et~al\mbox{.}}{2014}]%
        {shao:14}
\bibfield{author}{\bibinfo{person}{L. Shao}, \bibinfo{person}{R. Yan},
  \bibinfo{person}{X. Li}, {and} \bibinfo{person}{Y. Liu}.}
  \bibinfo{year}{2014}\natexlab{}.
\newblock \showarticletitle{From Heuristic Optimization to Dictionary Learning:
  A Review and Comprehensive Comparison of Image Denoising Algorithms}.
\newblock \bibinfo{journal}{\emph{IEEE Transactions on Cybernetics}}
  \bibinfo{volume}{44}, \bibinfo{number}{7} (\bibinfo{year}{2014}),
  \bibinfo{pages}{1001--1013}.
\newblock


\bibitem[\protect\citeauthoryear{Stead}{Stead}{2016}]%
        {p2020}
\bibfield{author}{\bibinfo{person}{R. Stead}.} \bibinfo{year}{2016}\natexlab{}.
\newblock \bibinfo{title}{{P2020} - Standard for Automotive System Image
  Quality}.
\newblock
  \bibinfo{howpublished}{\url{https://standards.ieee.org/develop/project/2020.html}}.
\newblock


\bibitem[\protect\citeauthoryear{Sun and Saenko}{Sun and Saenko}{2016}]%
        {sun2016deep}
\bibfield{author}{\bibinfo{person}{B. Sun} {and} \bibinfo{person}{K. Saenko}.}
  \bibinfo{year}{2016}\natexlab{}.
\newblock \showarticletitle{Deep coral: Correlation alignment for deep domain
  adaptation}. In \bibinfo{booktitle}{\emph{Computer Vision--ECCV 2016
  Workshops}}. Springer, \bibinfo{pages}{443--450}.
\newblock


\bibitem[\protect\citeauthoryear{Tang and Eliasmith}{Tang and
  Eliasmith}{2010}]%
        {TE:10}
\bibfield{author}{\bibinfo{person}{Y. Tang} {and} \bibinfo{person}{C.
  Eliasmith}.} \bibinfo{year}{2010}\natexlab{}.
\newblock \showarticletitle{Deep networks for robust visual recognition}. In
  \bibinfo{booktitle}{\emph{Proceedings of the International Conference on
  Machine Learning}}. \bibinfo{pages}{1055--1062}.
\newblock


\bibitem[\protect\citeauthoryear{Tang, Salakhutdinov, and Hinton}{Tang
  et~al\mbox{.}}{2012}]%
        {tang2012robust}
\bibfield{author}{\bibinfo{person}{Y. Tang}, \bibinfo{person}{R.
  Salakhutdinov}, {and} \bibinfo{person}{G. Hinton}.}
  \bibinfo{year}{2012}\natexlab{}.
\newblock \showarticletitle{Robust {B}oltzmann machines for recognition and
  denoising}. In \bibinfo{booktitle}{\emph{Proceedings of Computer Vision and
  Pattern Recognition}}. \bibinfo{pages}{2264--2271}.
\newblock


\bibitem[\protect\citeauthoryear{Tseng, Yu, Yang, Mannan, Arnaud,
  Nowrouzezahrai, Lalonde, and Heide}{Tseng et~al\mbox{.}}{2019}]%
        {tseng2019hyperparameter}
\bibfield{author}{\bibinfo{person}{Ethan Tseng}, \bibinfo{person}{Felix Yu},
  \bibinfo{person}{Yuting Yang}, \bibinfo{person}{Fahim Mannan},
  \bibinfo{person}{Karl~ST Arnaud}, \bibinfo{person}{Derek Nowrouzezahrai},
  \bibinfo{person}{Jean-Fran{\c{c}}ois Lalonde}, {and} \bibinfo{person}{Felix
  Heide}.} \bibinfo{year}{2019}\natexlab{}.
\newblock \showarticletitle{Hyperparameter optimization in black-box image
  processing using differentiable proxies}.
\newblock \bibinfo{journal}{\emph{ACM Transactions on Graphics (SIGGRAPH)}}
  \bibinfo{volume}{38}, \bibinfo{number}{4} (\bibinfo{year}{2019}),
  \bibinfo{pages}{27}.
\newblock


\bibitem[\protect\citeauthoryear{Tzeng, Hoffman, Saenko, and Darrell}{Tzeng
  et~al\mbox{.}}{2017}]%
        {tzeng2017adversarial}
\bibfield{author}{\bibinfo{person}{E. Tzeng}, \bibinfo{person}{J. Hoffman},
  \bibinfo{person}{K. Saenko}, {and} \bibinfo{person}{T. Darrell}.}
  \bibinfo{year}{2017}\natexlab{}.
\newblock \showarticletitle{Adversarial discriminative domain adaptation}.
\newblock \bibinfo{journal}{\emph{arXiv preprint arXiv:1702.05464}}
  (\bibinfo{year}{2017}).
\newblock


\bibitem[\protect\citeauthoryear{Tzeng, Hoffman, Zhang, Saenko, and
  Darrell}{Tzeng et~al\mbox{.}}{2014}]%
        {tzeng2014deep}
\bibfield{author}{\bibinfo{person}{E. Tzeng}, \bibinfo{person}{J. Hoffman},
  \bibinfo{person}{N. Zhang}, \bibinfo{person}{K. Saenko}, {and}
  \bibinfo{person}{T. Darrell}.} \bibinfo{year}{2014}\natexlab{}.
\newblock \showarticletitle{Deep domain confusion: Maximizing for domain
  invariance}.
\newblock \bibinfo{journal}{\emph{arXiv preprint arXiv:1412.3474}}
  (\bibinfo{year}{2014}).
\newblock


\bibitem[\protect\citeauthoryear{Vasiljevic, Chakrabarti, and
  Shakhnarovich}{Vasiljevic et~al\mbox{.}}{2016}]%
        {vasiljevic2016examining}
\bibfield{author}{\bibinfo{person}{I. Vasiljevic}, \bibinfo{person}{A.
  Chakrabarti}, {and} \bibinfo{person}{G. Shakhnarovich}.}
  \bibinfo{year}{2016}\natexlab{}.
\newblock \showarticletitle{Examining the Impact of Blur on Recognition by
  Convolutional Networks}.
\newblock \bibinfo{journal}{\emph{arXiv preprint arXiv:1611.05760}}
  (\bibinfo{year}{2016}).
\newblock


\bibitem[\protect\citeauthoryear{Yahiaoui, Hughes, Horgan, Deegan, Denny, and
  Yogamani}{Yahiaoui et~al\mbox{.}}{2019}]%
        {yahiaoui2019optimization}
\bibfield{author}{\bibinfo{person}{Lucie Yahiaoui}, \bibinfo{person}{Ciar{\'a}n
  Hughes}, \bibinfo{person}{Jonathan Horgan}, \bibinfo{person}{Brian Deegan},
  \bibinfo{person}{Patrick Denny}, {and} \bibinfo{person}{Senthil Yogamani}.}
  \bibinfo{year}{2019}\natexlab{}.
\newblock \showarticletitle{Optimization of ISP parameters for object detection
  algorithms}.
\newblock \bibinfo{journal}{\emph{Electronic Imaging}} \bibinfo{volume}{2019},
  \bibinfo{number}{15} (\bibinfo{year}{2019}), \bibinfo{pages}{44--1}.
\newblock


\bibitem[\protect\citeauthoryear{Zhang, Zuo, Chen, Meng, and Zhang}{Zhang
  et~al\mbox{.}}{2016}]%
        {zhang2016beyond}
\bibfield{author}{\bibinfo{person}{Kai Zhang}, \bibinfo{person}{Wangmeng Zuo},
  \bibinfo{person}{Yunjin Chen}, \bibinfo{person}{Deyu Meng}, {and}
  \bibinfo{person}{Lei Zhang}.} \bibinfo{year}{2016}\natexlab{}.
\newblock \showarticletitle{Beyond a {G}aussian Denoiser: Residual Learning of
  Deep {CNN} for Image Denoising}.
\newblock \bibinfo{journal}{\emph{arXiv preprint arXiv:1608.03981}}
  (\bibinfo{year}{2016}).
\newblock


\bibitem[\protect\citeauthoryear{Zhang, Wu, Buades, and Li}{Zhang
  et~al\mbox{.}}{2011}]%
        {zhang2011color}
\bibfield{author}{\bibinfo{person}{L. Zhang}, \bibinfo{person}{X. Wu},
  \bibinfo{person}{A. Buades}, {and} \bibinfo{person}{X. Li}.}
  \bibinfo{year}{2011}\natexlab{}.
\newblock \showarticletitle{Color demosaicking by local directional
  interpolation and nonlocal adaptive thresholding}.
\newblock \bibinfo{journal}{\emph{Journal of Electronic Imaging}}
  \bibinfo{volume}{20}, \bibinfo{number}{2} (\bibinfo{year}{2011}),
  \bibinfo{pages}{023016}.
\newblock


\end{thebibliography}


\end{document}